\documentclass[12pt,letterpaper]{article}

\usepackage{amsmath}
\usepackage{amssymb}
\usepackage{amsthm}
\usepackage{bm}
\usepackage{graphicx}
\usepackage{ifthen}
\usepackage{mathtools}
\usepackage{subcaption}
\usepackage[normalem]{ulem}

\renewcommand{\vec}[1]{\bm{#1}}

\newcommand{\abs}[1]{\left\lvert #1 \right\rvert}

\newcommand{\tr}{^{\mathsf{T}}}

\newcommand{\vecnorm}[1]{\left\| #1 \right\|}
\newcommand{\innerprod}[2]{\left<#1, #2\right>}

\newcommand{\yestag}{\stepcounter{equation}\tag{\theequation}}

\newcommand{\figref}[1]{\figurename~\ref{#1}}

\newcommand{\col}[1]{%
  \underset{\rule[-1.0em]{0.1ex}{1.2em}}{%
    \overset{\rule[0.2em]{0.1ex}{1.2em}}{#1}%
  }}

\newtheorem*{proposition}{Proposition}
\newtheorem*{dnprop}{Proposition (Decomposition of $\D\n$)}
\newtheorem*{d2nlemma}{Lemma (Decomposition of $\D^2\n$)}

\renewcommand{\k}{\kappa}
\renewcommand{\l}{\vec{\ell}}
\newcommand{\n}{\vec{n}}
\renewcommand{\b}{\vec{b}}
\newcommand{\nt}{\tilde\n}
\newcommand{\s}{\vec{s}}
\renewcommand{\v}{\vec{v}}
\newcommand{\w}{\vec{w}}
\newcommand{\x}{\vec{x}}
\newcommand{\y}{\vec{y}}

\newcommand{\zhat}{\hat{\vec{z}}}

\newcommand{\g}{\vec{g}}
\newcommand{\pdf}[1]{f_{#1}}

\newcommand{\U}[1]{%
  \if0#1 U\fi%
  \if1#1 Q\fi%
  \if2#1 B\fi%
}
\renewcommand{\S}[1]{%
  \if0#1 \Sigma\fi%
  \if1#1 S\fi%
  \if2#1 D\fi%
}
\newcommand{\V}[1]{%
  \if0#1 V\fi%
  \if1#1 R\fi%
  \if2#1 C\fi%
}
\newcommand{\Wt}{\tilde{W}}

\newcommand{\kf}{{\kappa_1}_s}
\newcommand{\kg}{{\kappa_1}_t}
\newcommand{\kh}{{\kappa_2}_s}
\newcommand{\ki}{{\kappa_2}_t}

\newcommand{\valpha}{\vec{\alpha}}
\newcommand{\vbeta}{\vec{\beta}}
\newcommand{\vgamma}{\vec{\gamma}}
\newcommand{\gradI}{\nabla\!I}

\newcommand{\calA}{\mathcal{A}}
\newcommand{\calK}{\mathcal{K}}
\newcommand{\calAunf}{\calA_{(1)}}
\newcommand{\D}{\mathcal{D}}
\renewcommand{\L}{\mathcal{L}}
\newcommand{\R}{\mathbb{R}}

\DeclareMathOperator{\vecop}{vec}
\DeclareMathOperator{\vechop}{vech}
\DeclareMathOperator{\trace}{tr}
\renewcommand{\det}{\operatorname{det}}

\newcommand{\pd}[2]{\frac{\partial#1}{\partial#2}}
\newcommand{\pdd}[2]{\frac{\partial^2#1}{\partial{#2}^2}}

\newcommand{\alignedarrow}{\!\!\Longrightarrow}

\begin{document}
\title{What's In A Patch, I: \\
Tensors, Differential Geometry and Statistical Shading Analysis}
\author{Daniel Niels Holtmann-Rice \\
Department of Computer Science\\
Yale University \and 
Benjamin S. Kunsberg \\
Department of Applied Mathematics \\
Brown University \\
\and
Steven W. Zucker\\
Depts. of Computer Science and Biomedical Engineering\\
Yale University}
\date{\today}

\maketitle

\begin{abstract}

We develop a linear algebraic framework for the shape-from-shading problem, because 
tensors arise when scalar (e.g. image) and vector (e.g. surface
normal) fields are differentiated multiple times. The work is in two
parts. In this first part we investigate when image derivatives exhibit invariance to
changing illumination by calculating the statistics of image
derivatives under general distributions on the light source. 
We computationally validate the hypothesis that image orientations (derivatives) provide increased invariance to illumination by showing (for a Lambertian model) that a shape-from-shading algorithm matching gradients instead of intensities provides more accurate reconstructions when illumination is incorrectly estimated under a flatness prior. 

\end{abstract}

\section{Introduction}

Shape-from-shading is a classical ill-posed problem, which requires additional structure (assumptions) to make it well-posed. The classical approach is based on solving partial differential equations or solving integral versions with different regularizers (priors). (A background review is provided in the next Section.) Instead of the relatively 'flat' model implied by a differential equation, that is, the relationship between derivatives of the same order across position, our approach considers the structure of increasing derivatives at the same position. Our motivation is to understand intuitively how differential structure in the image relates to differential structure on the surface, and is based on ideas from linear algebra and differential geometry.

Intuitively, for shape-from-shading, 
if one were to 'drill down' in derivatives for the surface then this should
correspond to analogous derivatives for the image.
Two classes of questions arise. First, working at similar levels of
differentiation, which image (derivatives) are most likely given
certain shape (normal) derivatives? Second, working across many levels
of derivatives in a 'Taylor' sense, which surfaces could correspond to
which derivatives? We shall address both of these questions in the
body of the paper. We consider the first of these questions in this
paper; the second is considered in the companion paper \cite{tensors-arxiv-2}.

\begin{figure}
 \begin{center}
   \begin{tabular}{c}
  \includegraphics[trim={0 8cm 0 3.5cm},clip, width = .7 \linewidth]{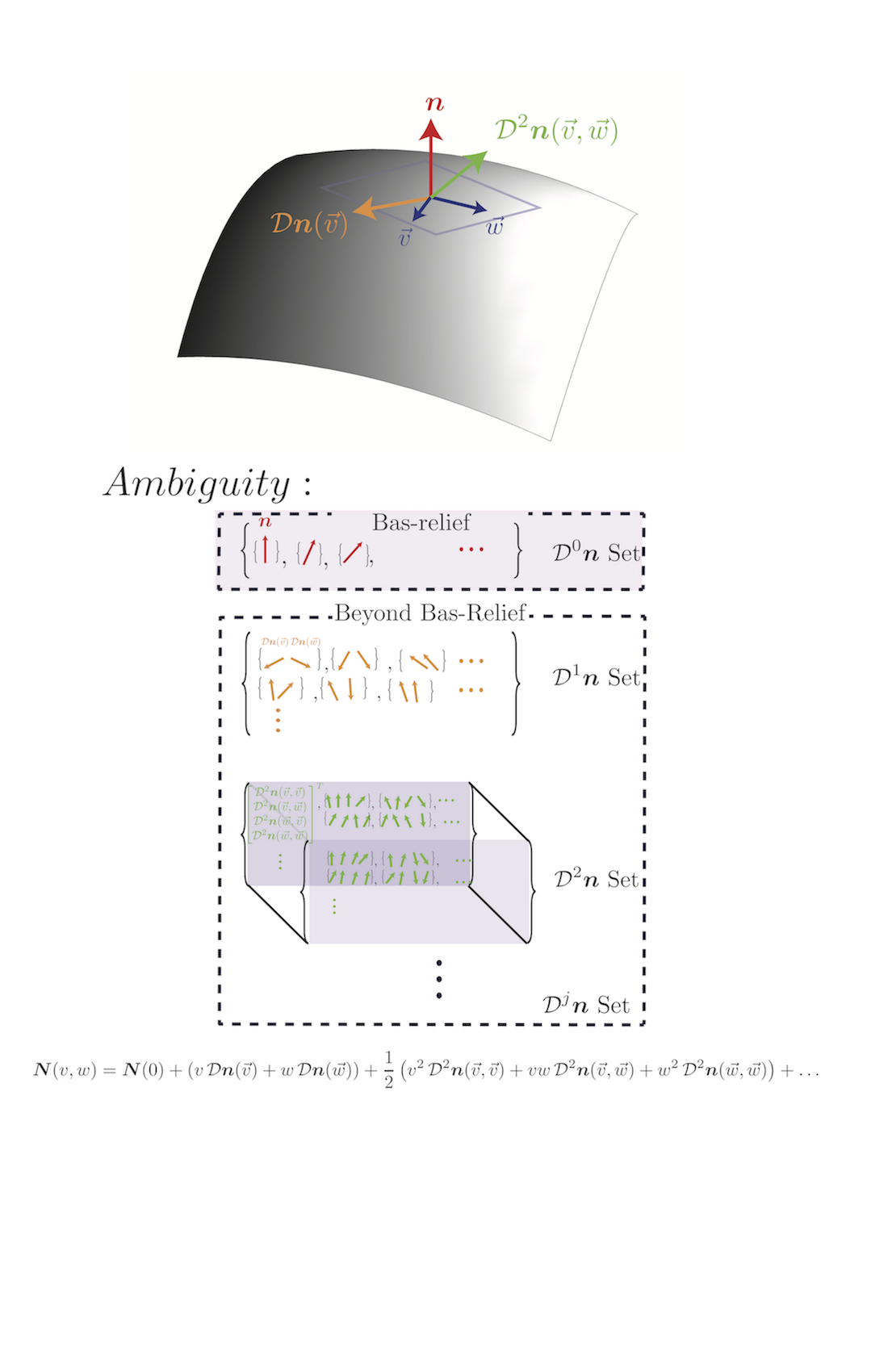}\\
(a) \\
  \includegraphics[width = .6 \linewidth]{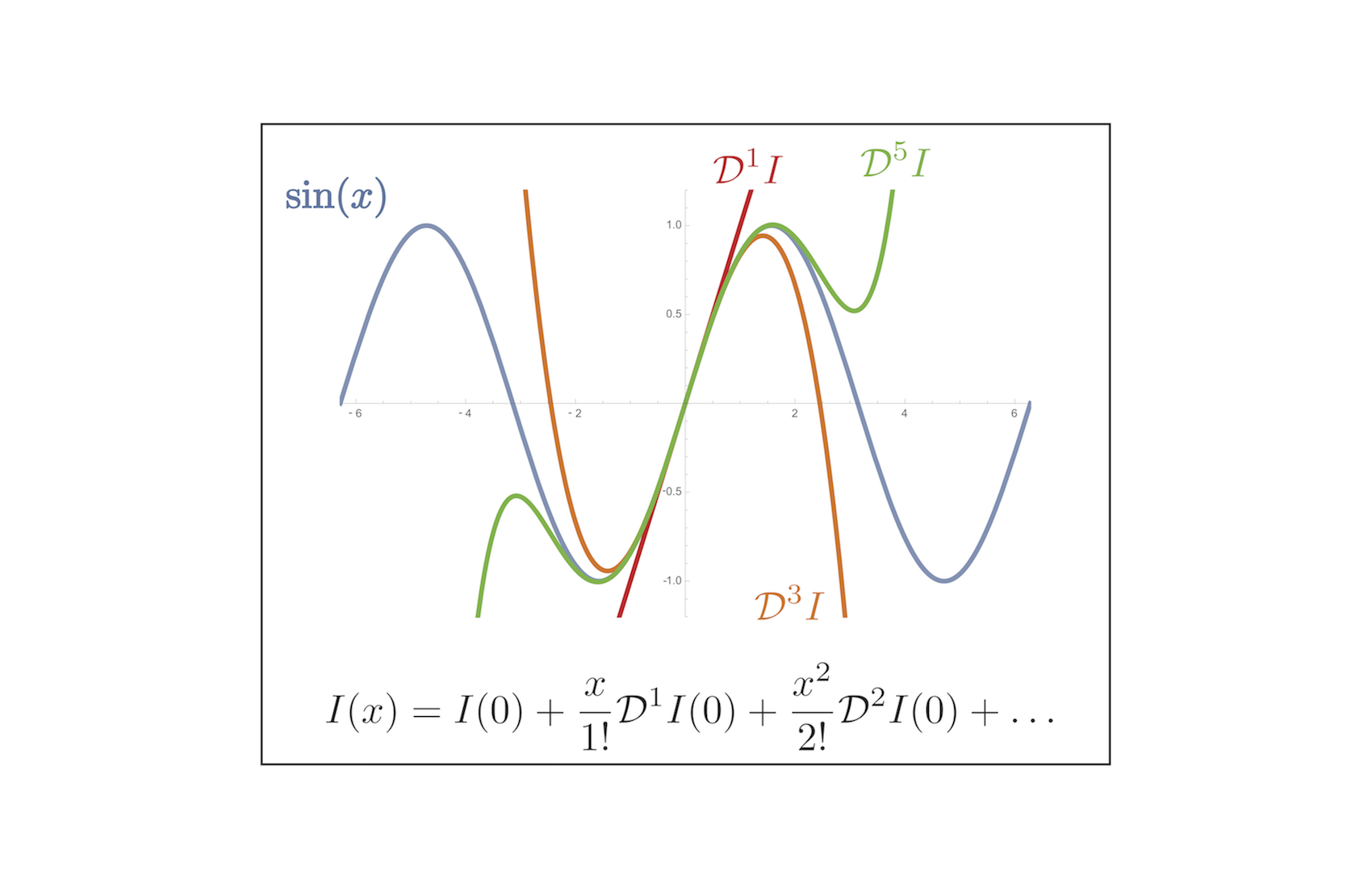}\\
(b) \\
    \end{tabular}
 \caption{(a) How tensors arise with increasing order of differentiation.
(a) The top figure shows the low-order geometry of the Lambertian
shape-from-shading problem on an image patch.
Below this are the sets of possibilities for each order of
differentiation, illustrating the figurative ambiguity involved.
Notice how the multi-linear tensor structure increases in complexity.
(b) Taylor's theorem in 1-D for a sinusoidal function, illustrating
how the domain of convergence (patch size) is related to order of derivative.}
 \end{center}
 \label{fig:taylora}
\end{figure}

Specifically, in the classical Lambertian shading model with a single, distant light source \cite{Horn:1970tf, Horn:1975tq}, the image intensity at a point is the inner product of the surface normal with the (typically unknown) light-source direction (also assuming orthographic projection). Note that this implies a scalar (image $I(x,y)$) field is related to a vector (surface normal $\n(x,y)$ ) field.  Applying the chain rule yields:

\begin{align*}
  I &= \l\tr \n + \beta\\
  \D I &= \l\tr \D\n \\
  \D^2 I & \leadsto \l\tr \D^2\n \\
  \D^3 I & \leadsto \l\tr \D^3\n \\
  &\cdots
\end{align*}
where, for clarity, dependence on image location is suppressed. 
Tensors arise naturally in this exercise, as the  representation of
derivatives (of derivatives ...) of a vector
(\figref{fig:taylora}). In particular, the derivative of the surface
normal, the shape operator $\D \n$, provides a measure of how the
normal changes if you move in a direction $\v$ (informally, a type of
directional curvature); this can be represented as a matrix
(the shape operator) applied to a vector. The next derivative, $\D^2 \n$
must be 'hit' by two vectors, which suggests that it is a 'matrix' of
'matrices,'  a much more complex object. Of course, working with
higher derivatives suggests a richer description of the patch, in the
sense of Taylor, which of course motivates a lot of our work.

For the image gradient, our analysis confirms the intuitive observation that cylindrical patches are the most likely surface patches (knowing only the first order structure of the image). Thus generic considerations arise, along with the associated algebraic notion of rank.

Once constructs such as image derivatives and Hessians arise, the question of which is 'most likely' follows immediately. Such questions are at the heart of machine learning and statistical approaches.  We employ the tensor machinery to derive the appropriate probability distributions for the first few derivatives. We provide explicit formulas for the image gradient and Hessian, conditioned on relevant surface parameters, under general light source distributions.  These distributions provide insight into which types of surfaces should be most invariant under different lighting (and other) conditions. 

For the image Hessian we examine when the matrix of second derivatives of the surface normal is rank 1, which restricts the space of associated image Hessians to lie along a line (vary only by a scaling factor) as the light source is varied. This is the basis for the assumption that the normal does not change in the isophote direction.  Despite the somewhat obvious nature of this ``prior'', it is relatively powerful, since it provides a specific constraint based on observable image features. 
In the process, we derive decompositions for $\D\n$ and $\D^2\n$, the first and second derivatives of the surface normal. These decompositions make explicit the dependence of these derivatives on the natural surface parameters of slant, tilt, and the principal curvatures and their derivatives, providing geometrically intuitive machinery for the invariance analysis. 

Finally, we exploit image derivatives 
in a Markov random field realization of a shape-from-shading algorithm. A flatness prior  deemphasizes the many possible surface variations as well as the skew from bas-relief. The statistical analysis suggests that working with differentials helps, and our experiments show that
the image gradient (or shading flow field \cite{Koenderink:1980bm, Breton:1996ic}) is more invariant to light source errors than similiar computations based on the raw intensities. 

The improved invariance from the image gradient corroberates results from several different areas. A number of results in human psychophysics, for example \cite{Fleming:2011dd, Fleming:2013gx}, highlight the role of orientations. In fact,  early processing in mammalian visual systems are essentially based on lifting the image into a position/orientation representation \cite{ben2004geometrical, sarti2008symplectic}. On the recognition side, 
convolutional neural networks \cite{2012arXiv1206.6445T} almost universally have an early stage of oriented filters.
In the end, we believe that a deeper appreciation of the rich connections between tensor analysis, differential geometry, and image structure can help to inform a new generation of approaches to the shape-from-shading and other vision problems. 

A note on reading this paper. We exploit the tensor structure relevant to relate image derivatives to normal field derivatives. 
To make this paper self-contained, we two Appendices with appropriate background material. 
The less experienced reader might glance at Appendix A, to get the basic notations for differential geometry, followed by Appendix B, essential ideas from tensor analysis. 
Otherwise, the technical content begins in Sec.~\ref{chapter:stats}.

\section{Background}

Starting with the classical work of Horn \cite{Horn:1970tf, Horn:1975tq, Horn:1977vn}, the 
shape-from-shading inference problem is formulated as a
differential equations with solutions sought along characteristic strips (but see \cite{Dieft81}).
Subsequently \cite{Ikeuchi:1981dq} developed a variational approach, representing surface orientation in stereographic coordinates to allow the incorporation of constraints from the object boundary, and enforcing smoothness via a penalty on squared derivatives of these coordinates. 
Closer in spirit to this work, Pentland \cite{Pentland:1984vy} analyzed the first and second derivatives of image intensity of a Lambertian shaded surface, demonstrating in particular that all parameters of the image formation process (including lighting) can be recovered locally when the surface is assumed to be spherical. 
He also \cite{Pentland:1990hv} linearized the reflectance function so that the resulting (linear) PDE could be solved via a spectral method.
\cite{Zheng:1991tg} provided an algorithm for estimating the illuminant direction from image statistics, as well as a shape-from-shading algorithm based on forcing the gradients of the reconstructed image to match the input image gradients. The algorithm is based on an energy minimization, and estimates surface heights and gradients simultaneously.
\cite{Worthington:1999um} provided an update procedure to allow the image irradiance equation to be a hard constraint---so that image intensities are always perfectly matched by those implied by the inferred normals (and known light source). This involves projecting the normals onto the cone of normals whose angle with the light source direction is consistent with the observed image intensities. They then investigated  different regularization constraints, including those based on curvature consistency \cite{Ferrie:1992vq}, and on matching the observed image gradients. Our simulations are in approximate agreement with his. Additional early work is reviewed in \cite{Zhang:1999wm}. For related psychophysical experiments, see \cite{Erens:1993vh, Liu:2004gm, Todd:1983vv}.

Prados employed viscosity solutions \cite{Prados:2006gu}, and in \cite{Prados:2005ca} showed that the problem becomes well-posed when the lighting model incorporates attenuation. 
More recently, \cite{Barron:2012tt} infer shape, illumination (a spherical harmonic lighting model), and albedo simultaneously using a Bayesian approach. They impose priors on illumination, albedo, and shape---the latter of which consist of an assumption of flatness (to counter bas-relief amibguities), boundary constraints, and low mean-curvature variation. This leads to an energy minimization (or likelihood maximization) which they solve using a standard quasi-Newton technique (L-BFGS). Our experiments use a model influenced by theirs, and our calculations provide additional support for it.

Recently papers emerged that are more consistent with our approach philosophy. 
\cite{Zoran:2014to} approached the problem of estimating shape and illumination (also using a spherical harmonic lighting model) by appealing to the \emph{generic viewpoint assumption} \cite{Freeman:1994br, Freeman:1996jc}---incorporating a prior based on ``genericity'' which favors solutions stable under slight changes in viewpoint or light source position. This was enforced via a penalty on image change under slight global rotations of the inferred object. They also require integrability, but do not require boundary constraints or additional priors. Nonetheless, they achieve results competitive with \cite{Barron:2012tt}.

Finally, a few  papers are explicitly based on a patch model. Conceptually, the idea is to solve for local patches individually and then 
``stitch" them together \cite{Xiong:2015hj,  Kunsberg:2014gua}; and see also \cite{Ecker:2010uh}.  Such approaches are possibly biologically relevant \cite{Connor08}.
Of course, this basic idea also underlies the pde approach, where regularizers of (typically low order) are introduced for posedness issues. Closer to this paper is
\cite{Ecker:2010uh}, who formulate the problem as solving a (large) system of polynomials using modern homotopy solvers. This is feasible for small images, involves (up to) quartic interactions,
and leads to exact recovery of all possible solutions. 

\section{Statistics of Lambertian Shading}
\label{chapter:stats}

Psychophysically image orientations exhibit significant invariance to changes in environment and isotropic surface markings for specular \cite{Fleming:2004bj} and textured \cite{Fleming:2011dd,  Garding:1993gz} surfaces. In shape-from-shading, a confounding variable is the direction of illumination. When do image orientations, or other low-order image derivative structure, exhibit invariance to illumination in shape-from-shading? What local surface structure is most likely to have generated observed low-order image structure?

To answer these questions, we now investigate the likelihood and invariance properties of low-order image derivatives of a shaded surface patch under generic lighting. We first examine the image gradient; then we extend our analysis to the image Hessian, and ask what third order surface structure makes the image Hessian invariant (up to scaling) to changes in illumination? 
In the process, we derive decompositions for $\D\n$ and $\D^2\n$, the first and second derivatives of the surface normal. 

To begin, we express derivatives of $\n$ in the standard basis of $\R^3$, to obtain tractable expressions of arbitrary orders of image derivatives under a Lambertian lighting model (See Appendix A for basic definitions and notation):
\begin{align*}
  I &= \l\tr \n + \beta\\
  \D I &= \l\tr \D\n \\
  \vecop(\D^2 I)\tr &= \l\tr \D^2\n_{(1)} \\
  \vecop(\D^3 I)\tr &= \l\tr \D^3\n_{(1)} \\
  &\cdots
\end{align*}
where dependence of $I$ and $\n$ on image location is suppressed for clarity. (The $\vecop$ and ${-}_{(1)}$ notation used for higher order derivatives are be covered in \ref{sec:disthess}.) We assume unit albedo, but not unit norm of $\l$. Thus the above model incorporates hemispheric as well as standard point-source Lambertian lighting (although ignoring rectification of image intensities).

The various $\D^j I$ are best viewed as $j^\textrm{th}$-order tensors (see Appendix), describing changes (in changes [in changes \ldots]) of image intensity in different directions. Expressed in the standard basis for the image plane, they form arrays containing partial derivatives of image intensity. In particular,
\begin{align*}
  \D I &= \gradI\tr = (I_x\; I_y) \\
\intertext{is the image gradient, while}
  \D^2I &= \begin{pmatrix}
    I_{xx} & I_{xy} \\
    I_{xy} & I_{yy}
  \end{pmatrix}
\end{align*}
is the image Hessian.

Assuming smoothness of the surface, $I_{xy} = I_{yx}$, which implies that the $\D^j I$ are symmetric: any permutation of the order of indices when accessing an entry in the multidimensional array yields the same value (e.g., $\D^3 I_{212} = \D^3 I_{221}$), and the order of inputs doesn't matter (e.g., $\D^3 I(\valpha, \vbeta, \vgamma) = \D^3 I(\vgamma, \valpha, \vbeta)$). Similar symmetry applies to the $\D^j\n$, although only to the second and higher modes, meaning only the indices in second or higher position can be permuted without any change in the value of the accessed entry. (This is simply because the first mode corresponds to the component of the normal being differentiated.)

In general, there is clearly no one-to-one correspondence between a particular collection of image derivatives and the surface that generated them (if only!)---many different combinations of $\n$, $\D\n$, $\D^2\n$ can yield the same combination of $I$, $\D I$, and $\D^2 I$, depending on the direction of illumination $\l$. However, not all combinations of surface derivatives that can generate a given image structure are equally likely. For instance, having observed only the image gradient at a point, both a locally spherical and locally cylindrical (with major axis orthogonal to the gradient) surface could have generated the observed structure---however, the locally cylindrical surface is in a sense more likely, because the image gradient direction is \emph{invariant} to changes in $\l$ for cylindrical surfaces, while an arbitrary gradient direction can be elicited from the spherical surface by varying $\l$.

\subsection{Distribution of the Image Gradient}

Making this intuition precise---quantifying the likelihood of different surfaces generating a given image structure---requires putting a probability distribution on $\l$. For instance, considering only $\gradI$ and $\D\n$ for the moment, given a specific $\D\n$ and a distribution on $\l$ yields $P(\gradI,\l\,|\,\D\n)$, the joint probability of a given image gradient/light source combination given specific normal variation. This is a delta function, since given the surface structure and the light source there is only one possible resulting image gradient. However, marginalizing out the light source ``nuisance parameter'' yields $P(\gradI\,|\,\D\n)$, the probability density of image gradients for given surface structure. This is also known as the likelihood $\L(\D\n\,|\,\gradI)$ of the surface structure given the image gradient.

To calculate this distribution, recall that linearly transforming a random variable $\x \in X \subset \R^n$ with density $\pdf{\x}$ by a (full-rank) matrix $A_{n \times n} : X \to Y \subset \R^n$ yields a random variable $\y = A\x$ whose distribution is given by
\begin{equation}
  \pdf{\y}(\y) = \frac{1}{\lvert\det A\rvert}\pdf{\x}(A^{-1}\y).
  \label{eq:pdflintrans}
\end{equation}

This is most easily seen by considering a small cube of ``probability mass'' at the point $\y = \y_0 \in Y$, and transforming back to $X$ via $\x_0 = A^{-1}\y_0$. The density at $\x_0$ imposed by $\pdf{\x}$ is then scaled by the (relative) transformed volume of the cube, which is given by $\lvert\det A^{-1}\rvert = \frac{1}{\lvert\det A\rvert}$. We can apply this to $\gradI$ to calculate $\pdf{\gradI}$. While $\D\n$ is not square, meaning it has no proper inverse, we can use the pseudo-inverse $\D\n^+$ instead, and replace $\det A$ with $\sqrt{\det \D\n\tr\D\n}$ (the product of the singular values of $\D\n$).

One definition of the pseudo-inverse is in terms of the SVD of $\D\n$: with $\D\n = \U1\S1\V1\tr$ ($\U1_{3 \times 2}$ with $\U1\tr\U1 = I$, $\V1_{2 \times 2}$ orthogonal, $\S1_{2 \times 2}$ diagonal), $\D\n^+ = \V1\S1^+\U1\tr$, where $\S1^+$ is formed by inverting the non-zero diagonal entries of $\S1$.

What is the geometric interpretation of the SVD of $\D\n$? The columns of $R$ (rows of $R\tr$) indicate the directions in the image in which the normal changes the most and least, for a unit step in the image. In other words, these are the directions of maximal and minimal \emph{view-dependent} curvature. The singular values in $\S1$ indicate the norm of these changes in normal. The columns of $Q$ correspond to the directions (in the tangent plane) of these maximal and minimal changes.

View-dependent curvature is a useful concept (see for example \cite{Judd:2007hm} for an application to generating line drawings), but it is useful to preserve the separate effects of foreshortening and curvature because a form based on intrinsic curvatures is easier to parameterize: $\D\n$ has six elements, but only five degrees of freedom (equivalent to the intrinsic parameters described below). In other words, we can't just pick any two orthogonal unit length vectors for $Q$ above (three parameters), any two choices for the singular values $\S1$, and some direction in the image for $T$---this may not lie on the appropriate five-dimensional manifold of ``valid'' $\D\n$'s.

For these reasons, we express $\D\n$ in terms of the natural parameters of first and second order surface structure: the slant $\sigma$ (degree of foreshortening), tilt $\tau$ (direction maximal of foreshortening), $\k_1$ and $\k_2$ (principal curvatures), and $\phi$, the angle (in the tangent plane, relative to the tilt direction) of maximum principal curvature.

\begin{dnprop}
  With $\D\s = U\Sigma V\tr$ representing the SVD of the differential of the surface parameterization $\D\s$, $W$ the matrix of principal curvature directions (expressed in the tilt basis), and $K$ the diagonal matrix of principal curvatures,
  \begin{align}
    \D\n &= UWKW\tr\Sigma V\tr \label{eq:dndecomp1} \\
    \D\n^+ &= \V0\S0^{-1} W K^{-1} W\tr U\tr
  \end{align}
\end{dnprop}

See appendix (\ref{sec:dndecomp}) for the derivation. Note that $U$, the $3\times 2$ matrix of left singular vectors of $\D\s$, consists of the $\R^3$ directions in the tangent plane of maximal and minimal slant, $\Sigma$ is the diagonal matrix $\left(\!\!\begin{smallmatrix}\frac{1}{\cos\sigma} & 0 \\ 0 & 1\end{smallmatrix}\!\right)$, and $V$ is a $2\times 2$ rotation matrix parameterized by $\tau$ (so its first column is the tilt direction in the image).

The above decomposition tells a small story about how change in the normal is calculated from a step in the image. From right to left in \eqref{eq:dndecomp1}, we follow an image vector as it is (1) represented in the tilt basis in the image by multiplication against $V\tr$; (2) projected onto the surface and represented in the basis formed by the tilt direction and its orthogonal (this just involves scaling the component of $\v$ in the tilt direction by $\frac{1}{\cos\sigma}$) by $\Sigma$; (3) represented in the principal curvature basis via transformation by $W\tr$; (4) scaled by the principal curvatures $K$ to yield the change in normal; (5) transformed back into the tilt basis; (6) expanded into $\R^3$ by $U$.

The pseudo-inverse can be seen as performing exactly the above operations in reverse (with $U\tr$ projecting into the tilt basis in the tangent plane).

\subsubsection{Distribution of {$\gradI$}}

To apply the formula for a linear transformation of a density function \eqref{eq:pdflintrans}, note that our expression is $\gradI = \l\tr\D\n$. We compute
\begin{align*}
  \gradI\tr\D\n^+
  &= \l\tr\D\n\,\D\n^+ \\
  &= \l\tr UU\tr \\
  &= \l\tr_t.
\end{align*}
$UU\tr$ performs projection into the tangent plane, so $\l_t$ is the tangential component of the light source $\l$. (We assume here that $\D\n$ is full rank.)

From \eqref{eq:pdflintrans} the density can then be written
\begin{equation}
  \pdf{\gradI|\D\n}(\gradI|\D\n) = \frac{1}{\sqrt{\det\D\n\tr\D\n}}\pdf{\l_t}(\gradI\tr \D\n^+),
  \label{eq:dipdfunexpanded}
\end{equation}
where $\pdf{\l_t}$ is the density of $\l_t$. Given a particular form for $\pdf{\l}$, we can calculate  $\pdf{\l_t}$, the corresponding density for the projected light source $\l_t$ (since we have knowledge of the tangent plane orientation provided by $\D\n$). If $\pdf{\l}$ is rotationally symmetric (i.e., uniform in the \emph{direction} of incoming light), $\pdf{\l_t}$ will depend only on the magnitude of $\l_t$, i.e., $\pdf{\l_t|\n}(\l_t|\n) = \pdf{\vecnorm{\l_t}}(\vecnorm{\l_t})$.

To arrive at an expression of \eqref{eq:dipdfunexpanded} in terms of the surface parameters $\sigma$, $\tau$, $\k_1$, and $\k_2$, we evaluate $\sqrt{\det\D\n\tr\D\n}$: \begin{align*}
  \sqrt{\det \D\n\tr\D\n}
  &=
  \sqrt{\det(\V0\S0 W K^2 W\tr \S0\V0\tr)} \\
  &=
  \sqrt{\det(\S0)^2\det(K)^2} \\
  &=
  \frac{\abs{\k_1\k_2}}{\cos\sigma} \\
  &=
  \frac{\abs{\k_G}}{\cos\sigma}, \label{eq:detdn} \yestag
\end{align*}
where $\k_G$ is the Gaussian curvature of the surface.
\begin{proposition}
  Using the above notation, the density for the image gradient, conditioned on $\D\n$ and given a corresponding distribution on the projected light source $\pdf{\l_t}$, has the natural parameter form
  \begin{equation}
    \pdf{\gradI|\D\n}(\gradI|\D\n) = \frac{\cos\sigma}{\abs{\k_G}}\pdf{\l_t}(\gradI\tr \D\n^+).
    \label{eq:dipdfgeneral}
  \end{equation}  
\end{proposition}

One concern is that this density becomes degenerate when $\D\n$ is rank 1---the likelihood of image gradients in the row-space of $\D\n$ becomes infinite. Theoretically, this can be dealt with by restricting our probability measure to this rowspace, something we don't pursue here. For computational purposes, this can be mitigated by adding noise to the image formation model.

We note that \cite{Chen:2000ba} also derives a distribution for the image gradient, consistent with the result here, in the specific case of normally distributed light sources and ignoring the effect of foreshortening.

As an example, consider the density on $\l$ given by the uniform distribution on the unit sphere, $\pdf{\l} = \frac{1}{4\pi}\delta(\vecnorm{\l} - 1)$, where $\delta$ is the Dirac delta distribution. Projection of this distribution onto a plane then yields 
\begin{align}
  \pdf{\gradI|\D\n}(\gradI|\D\n)
  &= \frac{\cos\sigma}{2\pi\abs{\k_G}\sqrt{1 - \gradI\tr\D\n^{\!+}{\D\n^{\!+}}{\!\tr}\gradI}} \notag\\
  &= \frac{\cos\sigma}{2\pi\abs{\k_G}\sqrt{1 - \gradI\tr\V0\S0^{-1} W K^{-2} W\tr \S0^{-1}\V0\tr\gradI}}
  \label{eq:dipdfuniform},
\end{align}
valid whenever $\k_G \ne 0$, $0 \le \sigma < \frac{\pi}{2}$.

It is useful to examine certain invariance properties of the image gradient.
$\pdf{\gradI|\D\n}$, the likelihood of $\D\n$, increases as $|\kappa_G|$ decreases (when $\pdf{\l_t}(\gradI\tr\D\n^+)$ can be held constant). This happens whenever one or both of the surface curvatures are sufficiently small; i.e., when the surface is close to cylindrical or planar. This implies, for non-zero gradients, curved cylinders (with axis orthogonal to the gradient) should be preferred as the likeliest local surface patches (if all we know is the first order image structure). This point is confirmed empirically in the final experimental section.

For cylinders, there is a one-dimensional space of possible gradients (i.e., the row space of $\D\n$, which determines the possible $\gradI$), or in other words, only the scale of the gradient (not the direction, up to sign) can vary as the light source is changed. This is intuitively clear, however this perspective (considering the dimension of the row space of $\D\n$) scales nicely to analyzing the image Hessian, which we address next.

\subsection{Distribution of the Image Hessian}
\label{sec:disthess}

We now seek to go ``up a level'' to calculate the distribution of the image Hessian, given \emph{third}-order surface structure $\D^2\n$? As with the gradient, the Hessian is linearly related to $\l$, but now through $\D^2\n$. Since $\D^2\n$ is a third-order tensor, we must consider what is the appropriate analog of the pseudo-inverse and product of singular values?

Using tools from linear algebra \ref{chapter:mathbackground}), we ``unfold'' the tensor into a matrix. For a third-order tensor, there are three possible unfoldings, achieved by laying out the columns, rows, or ``depths'' of the tensor side-by-side as column vectors in a matrix. These are referred to as the mode-1, mode-2, and mode-3 unfoldings, and for a tensor $\calA$ are denoted $\calAunf$, $\calA_{(2)}$, and $\calA_{(3)}$, respectively. In general, the mode-$i$ unfolding $\calA_{(i)}$ selects column vectors for the unfolded matrix by fixing all indices but the $i$-th in the tensor. The order in which column vectors are put into the unfolded matrix is for most purposes arbitrary, so long as a consistent convention is adopted.

We work exclusively with the mode-1 unfolding, since this preserves the mode (dimension) of the tensor responsible for interaction with the light source. For $\D^2\n$, which is naturally $3\times2\times2$, its unfolding $\D^2\n_{(1)}$ is $3\times4$. This gives the expression for the image Hessian
\begin{equation}
  \vecop(H)\tr = \vecop(\D^2 I)\tr = \l\tr \D^2\n_{(1)}
  \label{eq:hessvec}
\end{equation}
The left hand side of this equation requires use of the vectorization operator, taking a matrix and forming a column vector from its entries. In general, care should be taken to ensure this operation is compatible with the tensor unfolding operation, although here since $H$ and $\D^2\n$ are compatibly symmetric both row- and column-major approaches yield the same result.

A delicacy derives from the fact that the Hessian contains four elements, but has the constraint (assuming smoothness) that both mixed partial derivatives ($I_{xy}$ and $I_{yx}$) are equal. Its vectorization $\vecop(H)$ therefore lives on a three-dimensional subspace of $\R^4$, so the density for the Hessian defined on $\R^4$ is singular---all of the probability mass resides on a Lebesgue measure 0 subspace. Consequently, 
We only consider volume with respect to this three-dimensional subspace. An alternative to the full vectorization operation for symmetric matrices is the ``half-vectorization'' operator $\vechop(H)$, retaining only the three distinct elements of $H$ (dropping one of the redundant components from the Hessian). Making the right hand side of \eqref{eq:hessvec} compatible is then achieved by multiplication against the matrix \[
  L = \begin{pmatrix}
    1 & 0 & 0 \\
    0 & \frac{1}{2} & 0 \\
    0 & \frac{1}{2} & 0 \\
    0 & 0 & 1
  \end{pmatrix},
\]
giving
\begin{equation}
  \vechop(H)\tr = \l\tr\D^2\n_{(1)} L. \label{eq:hess}
\end{equation}
Note that $L^+$ gives the ``duplication'' matrix, such that $\vechop(H)\tr L^+ = \vecop(H)\tr$ (when $H$ is symmetric).

\begin{d2nlemma}
  Applying the above notation for the unfolding operation, $\D^2\n$ and $\D^2\n^+$ can be decomposed into ``natural parameter'' forms given by
  \begin{align}
    \D^2\n_{(1)} &= \U0_3 W_3 \calAunf(W\tr\S0\V0\tr)^{\otimes 2} \\
    \D^2\n^+_{(1)}
    &=
    (\V0\S0^{-1} W)^{\otimes 2}\calAunf^+W_3\tr\U0_3\tr,
  \end{align}
  where
  \begin{equation}
    \calAunf
    =
    \begin{pmatrix}
      f & g & g & h \\
      g & h & h & i \\
      \k_1^2 & 0 & 0 & \k_2^2
    \end{pmatrix},
  \end{equation}
  and $f = \kf$, $g = \kg$, $h = \kh$, $i = \ki$ are the partial derivatives of the principal curvatures (in the principal directions), $C^{\otimes 2} = C\otimes C$ is the Kronecker product of a matrix with itself, and $U_3$ and $W_3$ are orthogonal extensions of $U$ and $W$ to $3\times 3$ matrices.
\end{d2nlemma}
A derivation of the above decomposition is in the appendix, where we also provide a closed form expression for $\calAunf^+$. Note that $U_3$ adds the normal vector as a third column of $U$, while $W_3$ embeds $W$ in the upper left of a $3\times 3$ identity matrix. We denote partial derivatives in the first (maximal) and second (minimal) principal directions by ${-}_s$ and ${-}_t$, respectively.

This decomposition is similar to the one derived for $\D\n$, in that it consists of sending image vectors into the basis formed by the principle curvature directions in the tangent plane (the Kronecker product in the decomposition above does this for each of the two inputs to $\D^2\n$), calculating the change (or change in change) of the normal, and expanding/rotating back out into the standard basis for $\R^3$.

To use the decomposition in calculating $\pdf{H|\D\n,\D^2\n}$, we must calculate $|\!\det(\D^2\n_{(1)}\,L)|$, which using the decomposition is
\begin{align*}
  \det(\D^2\n_{(1)}\,L)
  &= \det\left(U_3 W_3 \calAunf(W\tr\Sigma V\tr)^{\otimes 2}\right) \\
  &= \det\left(\calAunf \Sigma^{\otimes 2}L\right)
\intertext{(by ignoring rotation matrices \cite{Knill:2014hp})}
  &= \det\left(\calAunf L \left(\!\begin{smallmatrix}
    \sec^2 \sigma & 0 & 0 \\
    0 & \sec\sigma & 0 \\
    0 & 0 & 1
  \end{smallmatrix}\!\right)\right) \\
  &= \det(\calAunf L)\sec^3 \sigma \\
  &= \frac{\kappa_1^2 \left(h^2-g i\right) + \kappa_2^2 \left(g^2-f h\right)}{\cos^3 \sigma}.
\end{align*}

\begin{proposition}
Using the above notation, the density of the Hessian (conditioned on third order knowledge of the surface $\s$) for a given distribution on light sources $\pdf{\l}$ has the natural parameter form
\begin{align}
  \pdf{H|\s}(H|\s)
  =
  \frac{%
    \cos^3 \sigma \cdot
    \pdf{\l}\left(\vecop(H)\tr(\V0\S0^{-1} W)^{\otimes 2}\calAunf^+W\tr\U0\tr\right)}%
  {\abs{\kappa_1^2 \left(h^2-g i\right) + \kappa_2^2 \left(g^2-f h\right)}}.
\end{align}  
\end{proposition}

\subsection{Invariance Properties of the Image Hessian}

Under what circumstances does the image Hessian possess invariance to changes in light position? In particular, for cylindrical surfaces, the gradient is restricted to lie along a one-dimensional subspace---what are the analogs for third-order shape, i.e., where the Hessian is restricted to a one-dimensional subspace?

The decomposition derived for $\D^2\n_{(1)}$ affords an approach to answering this question. Recall
\begin{equation*}
  \vecop(H)\tr = \l\tr\D^2\n_{(1)} = \l\tr U_3W_3\calAunf(W\tr\Sigma V\tr)^{\otimes 2}.
\end{equation*}
Note that the rowspace of $\D^2\n_{(1)}$ spans the space of possible image Hessians for a given $\D^2\n$. Since $\D^2\n_{(1)}$ is $3\times 4$, whenever $\D^2\n$ is full (row) rank, the space of possible image Hessians is three-dimensional, i.e.\ \emph{any} possible image Hessian can be generated by positioning the light source appropriately. The space of possible Hessians is restricted only when $\D^2\n$ has reduced rank (one or more of its singular values is 0).

We now examine when $\D^2\n_{(1)}$ is rank 1. Since $U_3$, $W_3$, $V$, and $\Sigma$ are always full rank, this occurs when $\calAunf$ is rank 1. This in turn occurs when the rows or columns of $\calAunf$ are all scalar multiples of one another. The distinct columns of $\calAunf$ are \[
  \v_1 = \begin{pmatrix}
    f \\ g \\ \k_1^2
  \end{pmatrix}
  \qquad
  \v_2 = \begin{pmatrix}
    g \\ h \\ 0
  \end{pmatrix}
  \qquad
  \v_3 = \begin{pmatrix}
    h \\ i \\ \k_2^2
  \end{pmatrix}
\]
When the columns are scalar multiples of one another, then $\v_1 = \alpha \v_2 = \beta \v_3$. We can see immediately $\alpha = 0$, since $\alpha \k_1^2 = {v_2}_3 = 0$ (assuming $\k_1 \ne 0$). Consequently, $g = \alpha f = 0$, and $h = \alpha g = 0$. We're left with \[
  \v_1 = \begin{pmatrix}
    f \\ 0 \\ \k_1^2
  \end{pmatrix}
  \qquad
  \v_2 = \vec{0}
  \qquad
  \v_3 = \begin{pmatrix}
    0 \\ i \\ \k_2^2
  \end{pmatrix}.
\]
Three possibilities remain, distinguished by whether $\k_1$ and $\k_2$ are both zero, only $\k_1$ is non-zero, or both are non-zero (we assume $\k_1^2 \ge \k_2^2$ via our choice of basis).

\begin{proposition}
  The three circumstances under which $\D^2\n_{(1)}$ is rank 1 are
  \begin{enumerate}
    \item $\k_1 = \k_2 = 0$, and $(f,g)$, $(g,h)$ and $(h,i)$ all lie along the same line.

    \item $f = \kf \ne 0$ and $\beta = i = 0$, so $\v_3 = \vec{0}$ and $\k_2^2 = 0$.

    \item $f = g = h = i = 0$ but $\beta \ne 0$, and $\k_2^2 = \beta \k_1^2$.
  \end{enumerate}
\end{proposition}

{\em Case 1} above occurs when the surface has no curvature (locally planar or an inflection point of the surface normal). For this condition, the image gradient will always be 0, since there is no normal change in any direction, i.e., we are at a singular point in the image. Furthermore, since the Jacobian of the principal curvatures is given by $\left(\!\begin{smallmatrix}f & g \\ h & i\end{smallmatrix}\!\right) = \left(\!\begin{smallmatrix}\kf & \kg \\ \kh & \ki\end{smallmatrix}\!\right)$, and $(f,g)$ and $(h,i)$ are collinear, the principal curvatures are only changing in one direction (and there is another direction in which both principal curvatures remain 0). This occurs for example at the inflection point along a sigmoidal shape extruded along a straight line.

{\em Case 2} corresponds to a generalization of the cylinder---normal change (and change in normal change) occurs in only one direction. A corollary of this condition is that the image Hessian is itself rank 1. To see this, note that $\l\tr U_3W_3\calAunf = (a,0,0,0) = \vecop(X)$ for some $a$, letting $X$ be the matricization (inverse of the vectorization) of $(a,0,0,0)$. Then, with $M = W\tr\Sigma V\tr$ we have
\begin{equation}
  \vecop(H)\tr =\vecop(X)\tr M^{\otimes 2} = \vecop(M\tr X M),
\end{equation}
via an identity of the Kronecker product. Thus $H = M\tr X M$. Since $X$ is clearly rank 1, so is $H$.

This condition dictates that when the image Hessian is rank 1 and the gradient direction is orthogonal to the nullspace of the Hessian (intensity change and change in the gradient lie in the same direction), cylindrical solutions should be preferred. Or, in other words, we should assume the normal isn't changing in the isophote direction. Despite the somewhat obvious nature of this ``prior'', it is relatively powerful, since it provides a specific constraint based on observable image features.

In {\em Case 3} there is no third-order change at all. Because the third-order terms are exactly the derivatives of the principal curvatures, this means we are at a critical point of the principal curvatures---for example a local maximum or minimum, or a region of locally constant curvatures. This case reveals that the Hessian changes only up to an overall scaling factor (meaning properties like its eigenvectors and the ratio of its eigenvalues are preserved) under geometrically (and perceptually) interesting locations---namely, at extrema of curvature (such as often occur at the top/bottom of many bumps/dimples), or regions of constant curvature. (A simple example of the latter condition is of course the sphere, which has constant positive curvatures.) Furthermore, while the space of possible Hessians is one-dimensional in this situation, the Hessian itself will generally \emph{not} be rank 1.

\begin{figure}
  \centering
  \includegraphics[width=0.45\textwidth]{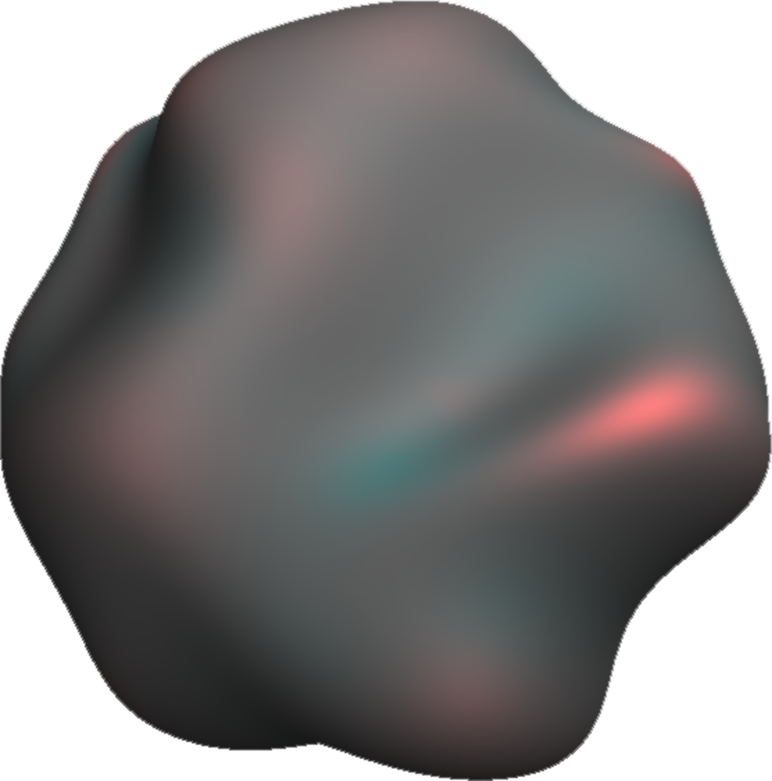}
  \hfill
  \includegraphics[width=0.45\textwidth]{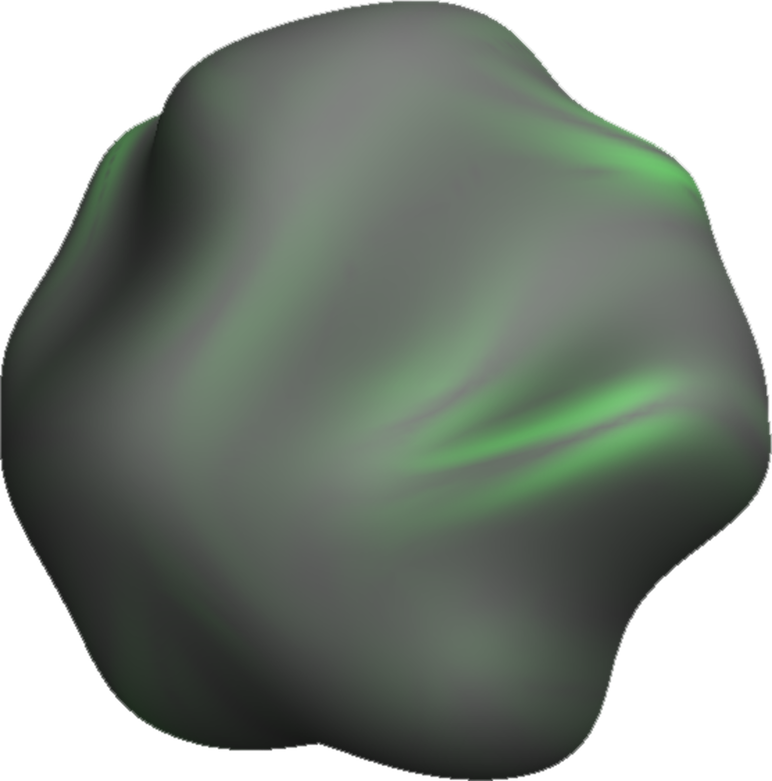}
  \caption[Visualization of 2nd and 3rd order structure]{Visualization of second and third order structure on a mesh. On the left, red and cyan indicate positive and negative Gaussian curvature regions, respectively. On the right, intensity of green indicates the maximum absolute value of the third order coefficients. An online interactive demo of this visualization is available at {http://dhr.github.io/mesh-curvatures}. The curvatures and third order terms are calculated via \cite{Rusinkiewicz:2004ia}.}
\end{figure}



\subsection{Combining the Gradient and Hessian}

Thus far, we have considered the gradient and Hessian separately, however
they are not independent: given knowledge of the surface $\s$, observing $\gradI$ provides information about the position of the light source, i.e., $\pdf{\l|\gradI,\D\n} \ne \pdf{\l}$. For instance, knowledge of a full-rank $\D\n$ and its accompanying observed image gradient restricts the light source to lie along a one-dimensional subspace parallel to the normal, since we can reconstruct the tangential component of the light source via $\l_t\tr = \gradI\tr\D\n^+$. Incorporating this effect yields an expression for the joint density. Expanding the joint distribution via the chain rule gives \begin{equation*}
  \pdf{\gradI,H|\s}
  =
  \pdf{H|\gradI,\s} \cdot \pdf{\gradI|\s}.
\end{equation*}
We have already calculated $\pdf{\gradI|\s}$ above, and $\pdf{H|\gradI,\s}$ depends on $\gradI$ only through the constrained distribution on $\l$. Thus \begin{equation}
  \pdf{\gradI,H|\s}
  =
  \frac{(\cos \sigma)^4}{\abs{\k_G}\abs{m}}
  \pdf{\l|\gradI,\s}(\vecop(H)\tr\D^2\n^+)\,\pdf{\l_t|\s}(\gradI\tr\D\n^+),
\end{equation}
where $m = \kappa_1^2 \left(h^2-g i\right) + \kappa_2^2 \left(g^2-f h\right)$. 

A joint density involving the image intensity is also possible via a similar approach---we note that in this case, $\pdf{\l|\gradI,I,\s}$ is in fact a delta function (knowing the intensity, gradient, the normal and its derivative, we can generically recover the light source). To expand,
under the Lambertian model, note that $I$ provides the component of the light source lying in the normal direction $\n$: $\l_n = \l\tr\n\n\tr = I\n\tr$ is the projection of the light source onto the normal. Additionally, note that $\D\n\D\n^+ = UU\tr$, which is the projection operator into the tangent plane. Thus $\gradI\tr\D\n^+ = \l\tr\D\n\D\n^+ = \l\tr U U\tr = \l_t$, the projection of $\l$ into the tangent plane. This gives the relation (for non-zero curvatures) $\l\tr = \l_n\tr + \l_t\tr = I \n\tr + \gradI\tr\D\n^+$. Substituting this in to the equation for the Hesssian gives
\begin{equation}
  \vecop(H)\tr = (I \n\tr + \gradI\D\n^+)\D^2\n_{(1)}.
\end{equation}
This is a linear algebraic formulation of the ``second order shading equations'' derived in \cite{Kunsberg:2014gua}.

When $\D^2\n_{(1)}$ is full rank, we can additionally express the light source via $\l\tr = \vecop(H)\tr\D^2\n_{(1)}^+$, which can be plugged into the formula for the image gradient and intensity to yield alternate expressions.

\subsection{Connections to Other Work}

Recall that we have
\begin{align*}
  \vecop(H)\tr &= \l\tr U_3W_3\calAunf(W\tr\Sigma V\tr)^{\otimes 2} \\
  &\alignedarrow \\
  \vecop(H)\tr(V\Sigma^{-1})^{\otimes 2}
  &=
  \l\tr U_3 W_3 \left(\!\begin{smallmatrix}
    f & g \,&\, g & h \\
    g & h \,&\, h & i \\
    \k_1^2 & 0 \,&\, 0 & \k_2^2
  \end{smallmatrix}\!\right)W^{\otimes 2}
\end{align*}
For any fixed choice of normal (fixing $U$, $\Sigma$, and $V$), light source, and third order terms, we can match a given image Hessian by specifying the principal curvatures and directions. There are four choices in general, corresponding to the choices of signs of the principal curvatures (as these get squared in $\calA$). A related result is called the ``four-fold ambiguity'' in \cite{Kunsberg:2014gua}. Subsequent work in \cite{Xiong:2015hj} used this observation in service of a shape-from-shading algorithm (assuming known light source). This work assumed third-order coefficients resulting from a Monge-patch expansion from the image plane were small---but note that these are different third order coefficients from those in $\calA$ (which are defined from the tangent plane). Third order coefficients defined from the image plane are view dependent, while those in $\calA$ are not (meaning they are invariant to rotations of the surface).

The approach described so far is similar to the notion of genericity described in \cite{Freeman:1994br}. That work provides a general derivation of ``generic'' (stability) priors in inference problems, using a Laplace approximation to derive a form for the posterior distribution of scene parameters (here, shape) given image data, by marginalizing out ``nuisance'' parameters that don't need to be precisely estimated (the light source). Due to our formulation, we have calculated the posterior exactly in the case of local Lambertian shading.


\section{Gradient-Based Shape-from-Shading}
\label{chapter:comp}

We now perform a computational experiment based on the previous analyses. 
We adopt a Markov random field (MRF) framework for structuring the inference \cite{Wang:2013fq}, so that effect of key points in the previous analysis can be evaluated. Specifically, from the
statistical computations we introduce a cylindricity potential, and second we suppress the variation in possible surface inferences by a flatness potential.
In the end, we show that matching image gradients is less sensitive to errors in assumed light source position than a reconstruction based on matching intensities directly.

We use an image triangulation as the base graph, and the gradient of surface depth as the latent variables.
We optimize an energy functional consisting of standard and non-standard terms.
\begin{itemize}
  \item Standard Terms:
\begin{itemize}
  \item Image intensities $  \phi_I$: via squared error; equivalent to assuming corruption by additive Gaussian noise. This unary  potential applies independently to each node $i$ in the triangulation:
We assume hemispheric lighting, to avoid large black regions in the image under oblique lighting conditions (when the normal faces away from the light source).
  \item Integrability  $ \phi_{\textrm{int}}$: penalizes deviation from symmetry of the estimated surface Hessian:
  \item Boundary  $\phi_\text{b}$: enforces orthogonality of estimated normals to the surface boundary.
\end{itemize}
\item Non-standard Terms:
\begin{itemize}
    \item Flatness: counters the bas-relief family \cite{Belhumeur:1999hd}; also used in \cite{Barron:2012tt};
            \item Image gradient
              \item Cylindricity: encourages normal change to happen in the direction of the image gradient by
penalizing normal change occuring in the isophote direction.
\end{itemize}
\end{itemize}

The energy functional is a simple summation:

\begin{equation}
  E(\g) = w_I \phi_I(\g) + w_{\gradI} \phi_{\gradI}(\g) + w_\text{int} \phi_\text{int}(\g) + w_\text{flat}\phi_\text{flat}(\g) + w_\text{cyl}\phi_\text{cyl}(\g).
\end{equation}

\noindent
and it is optimized using L-BFGS (Limited-memory Broyden-Fletcher-Goldfarb-Shanno) \cite{Liua:1989vk}.

\begin{table}[t!]
  \begin{tabular}{ccc}
    \hline
    Image & Ground Truth & Reconstruction \\
    \hline \\
    \parbox[c]{0.3\textwidth}{\includegraphics[width=0.3\textwidth]{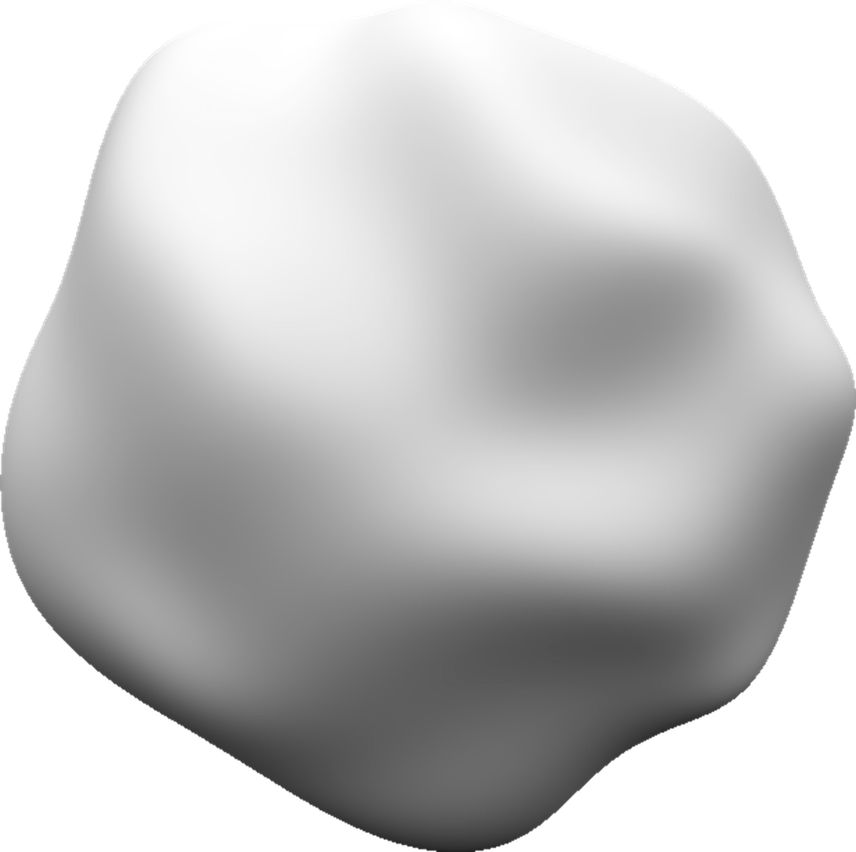}}
    &
    \parbox[c]{0.3\textwidth}{\includegraphics[width=0.3\textwidth]{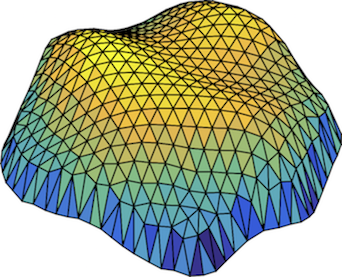}}
    &
    \parbox[c]{0.3\textwidth}{\includegraphics[width=0.3\textwidth]{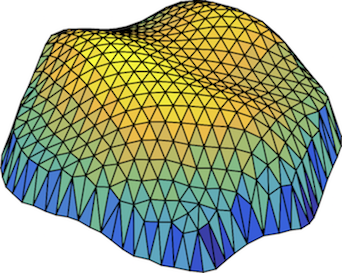}} \\
    \parbox[c]{0.3\textwidth}{\includegraphics[width=0.3\textwidth]{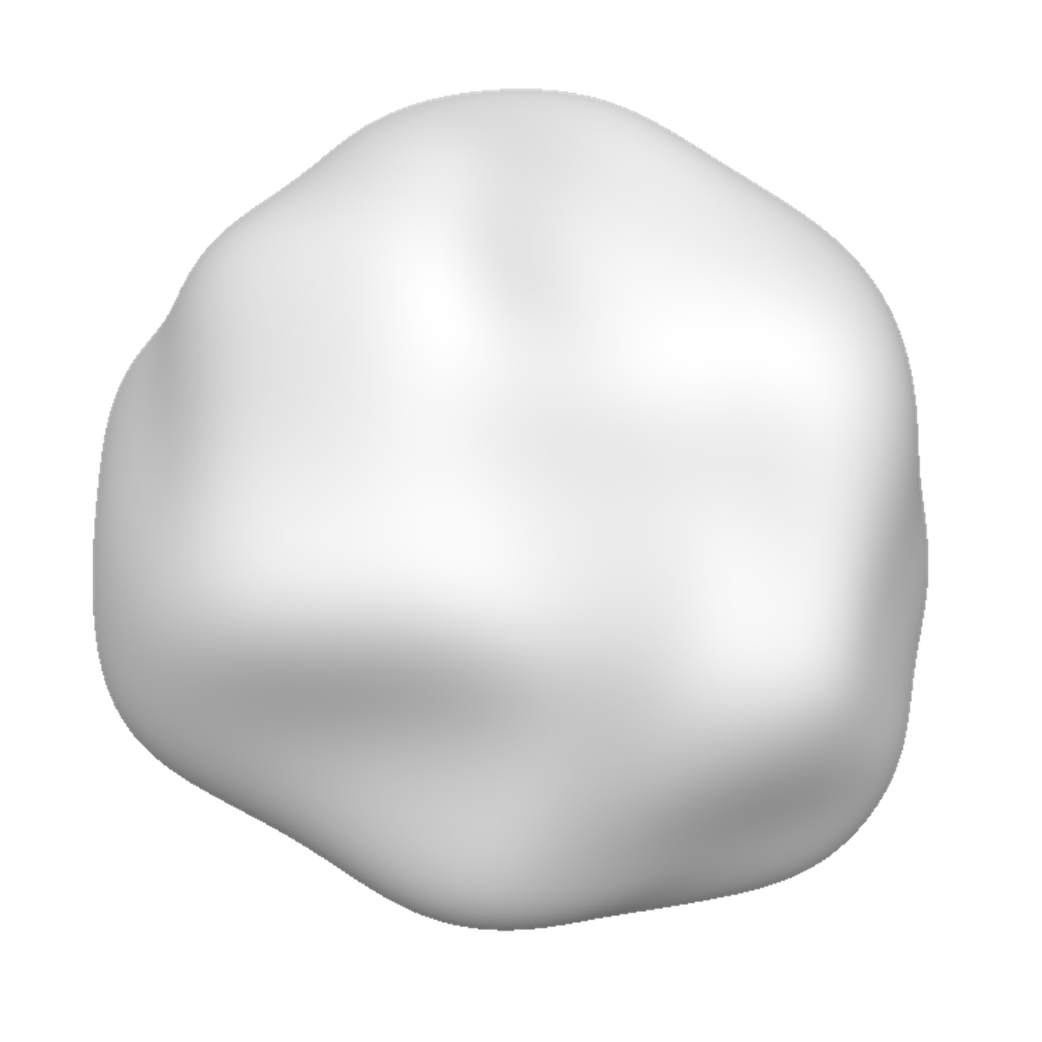}}
    &
    \parbox[c]{0.3\textwidth}{\includegraphics[width=0.3\textwidth]{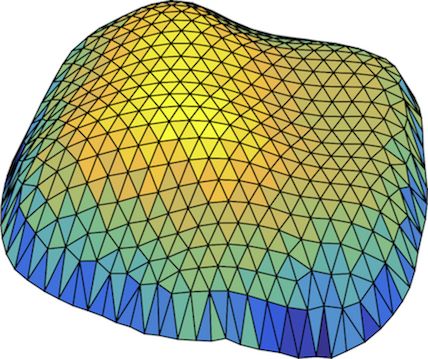}}
    &
    \parbox[c]{0.3\textwidth}{\includegraphics[width=0.3\textwidth]{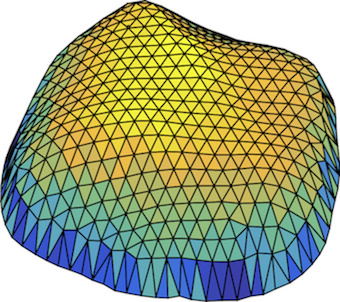}} \\
    \parbox[c]{0.3\textwidth}{\includegraphics[width=0.3\textwidth]{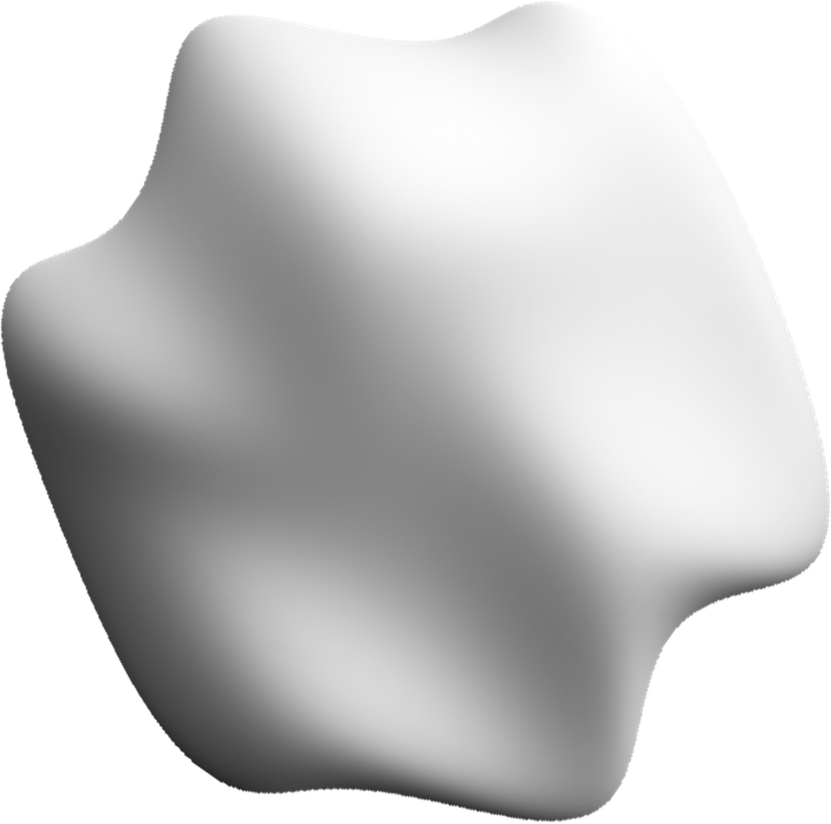}}
    &
    \parbox[c]{0.3\textwidth}{\includegraphics[width=0.3\textwidth]{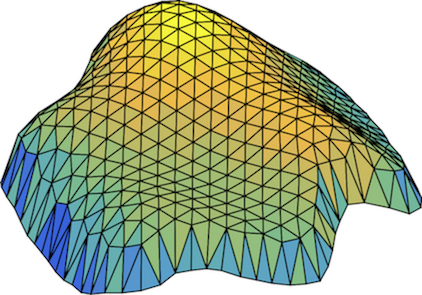}}
    &
    \parbox[c]{0.3\textwidth}{\includegraphics[width=0.3\textwidth]{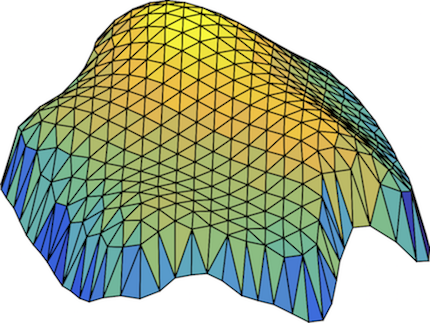}}
  \end{tabular}
  \caption{Example reconstructions, known light source. Mean/median angular errors in the reconstructed normals are, from top to bottom, 4.3/3.8, 6.8/3.6, 4.7/4.0 (in degrees).}
  \label{fig:reconstructions}
\end{table}

In the first experiment we demonstrate reconstructions based on intensities a known light source, for several shapes in Table \ref{fig:reconstructions}. The weights used are $w_I = 4$, $w_{\gradI} = 0$, $w_\text{int} = 150$, $w_\text{flat} = 0.001$. They were chosen to balance the magnitude of the contribution of each active term to the overall objective function value, which yielded good performance.

\begin{figure}
  \centering
  \newbox\figa
  \setbox\figa\hbox
    {\includegraphics[width=0.45\textwidth]{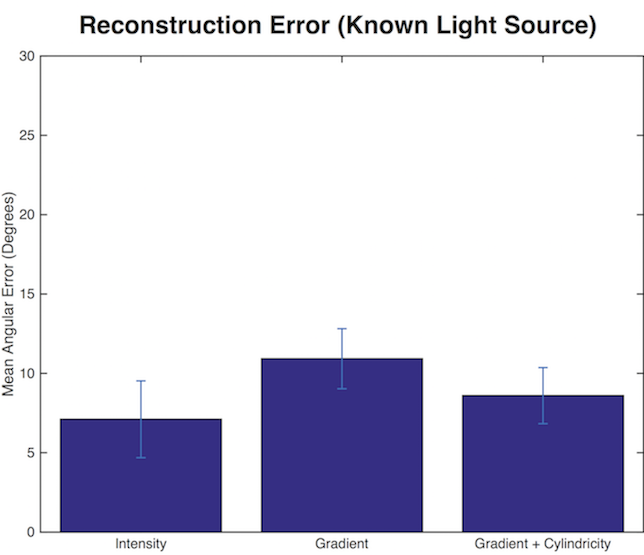}}
  \subcaptionbox{\label{fig:recknown}}{\unhcopy\figa}
  \hfill
  \subcaptionbox{\label{fig:recoffset}}
    {\includegraphics[height=\ht\figa]{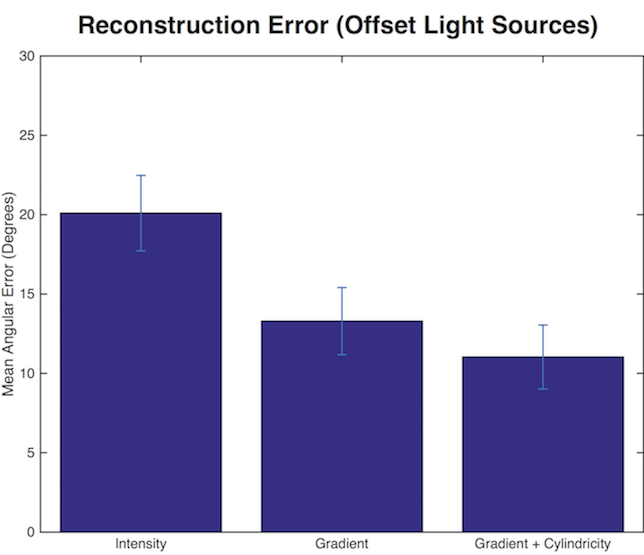}}
  \caption[Reconstruction errors under known and perturbed light sources]{(a) Reconstruction error under known light sources, averaged across shapes and lighting conditions. (b) Average reconstruction error under perturbed light sources.}
  \label{fig:recerrors}
\end{figure}

We empirically test our hypothesis that matching image gradients yields improved invariance to light source position when the assumed light source contains estimation error. To evaluate performance, we used a set of smooth but structured shapes, and seven initial light source positions. For each light source position, we perturb the source by 22.5 degrees in each of four directions (towards the viewer, away from the viewer, and clockwise and counterclockwise). We note that human observers frequently make errors of this magnitude in estimating the direction of illumination \cite{OShea:2010cp}.

We inferred shapes using three settings of the weights for the energy function $E$ above. First, we reconstructed based on image intensities ($w_I = 4$, $w_{\gradI} = 0$). A second reconstruction was performed using image gradients ($w_I = 0$, $w_{\gradI} = 100$). The weight for the gradient term was chosen so that performance was good on known light source images and so that its contribution to the overall energy was similar in magnitude to the contribution from the intensity term. Finally, a third reconstruction ($w_I = 0$, $w_{\gradI} = 100$, $w_\text{cyl} = 10$) was performed, including the cylindricity constraint. All reconstructions shared $w_\text{b} = 0.05$, $w_\text{int} = 150$, and $w_\text{flat} = 0.001$.

\begin{figure}
  \centering
  \includegraphics[width=0.95\textwidth]{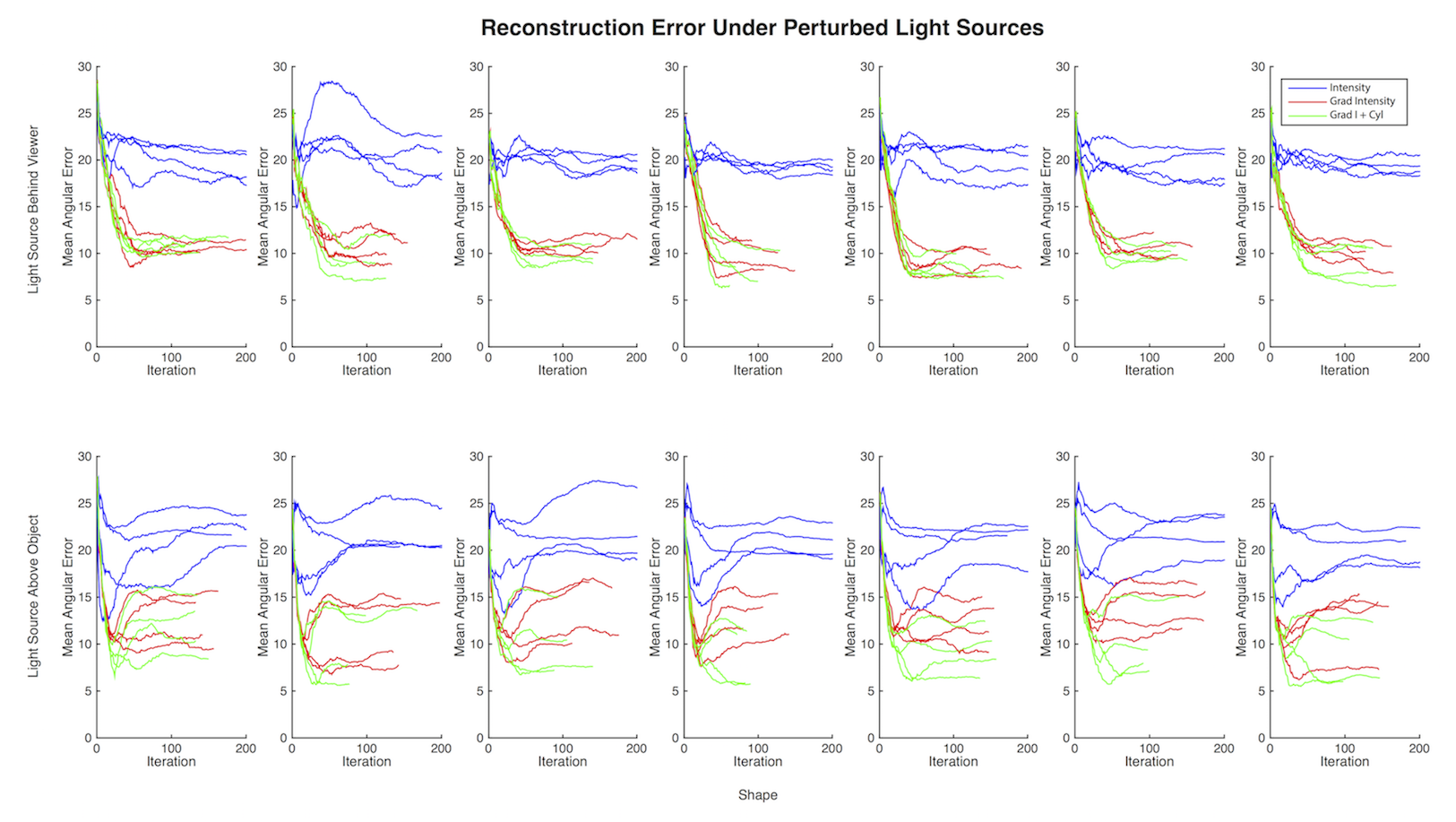}
  \caption[Time course of error under perturbed light sources]{Mean angular error of reconstructions under perturbed light sources. The initial position of light sources is close to behind the viewer in the top row, and close to above the object in the bottom row. Each same-colored line represents perturbation of the light source by 22.5 degrees in one of four directions. Shape varies by column. Blue lines represent reconstruction based on intensities; red lines based on gradients; green lines based on gradients plus a cylindricity potential. Note that error when reconstructing from gradients is substantially lower than when reconstructing from intensities.}
  \label{fig:recerriters}
\end{figure}

In \figref{fig:recerrors}, we show the mean angular error of reconstructions from intensities, gradients, and gradients plus the cylindricity term, for both exact and perturbed light sources, averaged across all shapes and lighting conditions. In \figref{fig:recerriters}, we plot the mean angular error (average angular difference between inferred and true normals) in the reconstruction for all shapes and light source positions, as a function of number of iterations of the optimization. Note that matching based on gradients tends to give both faster convergence and lower overall error. Some example reconstructions, comparing results from reconstructions from intensities to those from gradients, can be seen in \ref{fig:pertrecs}.

Adding the cylindricity constraint improves results further. While the results of \ref{chapter:stats} suggest assuming cylindricity when the gradient structure is locally parallel, this is not enforced explicitly in the constraint we employ here. However, a similar effect occurs as a byproduct of the optimization process (at the cost of some additional flatness in doubly curved regions)---satisfying both cylindricity and gradient matching penalties cannot be fully achieved when the gradient structure is curved (image Hessian full rank), however both can be fully satisfied in cylindrical regions.

We conclude that when the direction of illumination contains moderate error, image gradients provide a better target for ``matching'' based shape-from-shading algorithms.

\begin{table}[p!]
  \begin{tabular}{cccc}
    \hline
    Image & Ground Truth & Intensities & Gradients \\
    \hline \\
    \parbox[c]{0.22\textwidth}{\includegraphics[width=0.22\textwidth]{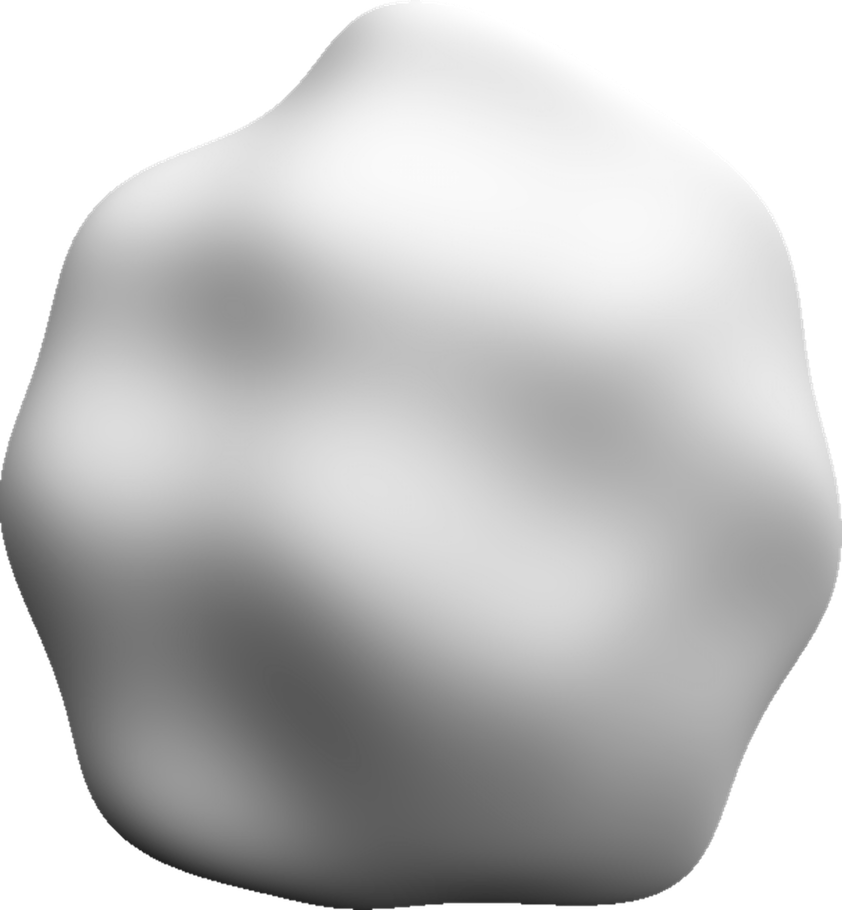}}
    &
    \parbox[c]{0.22\textwidth}{\includegraphics[width=0.22\textwidth]{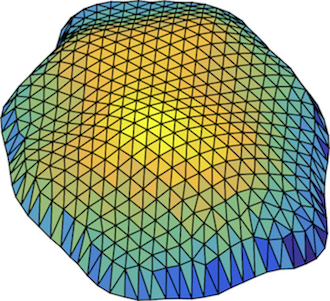}}
    &
    \parbox[c]{0.22\textwidth}{\includegraphics[width=0.22\textwidth]{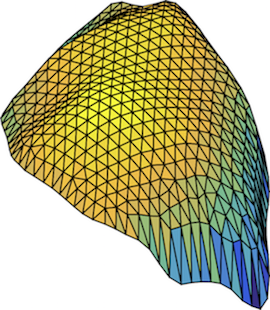}}
    &
    \parbox[c]{0.22\textwidth}{\includegraphics[width=0.22\textwidth]{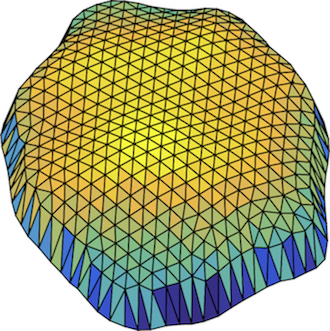}} \\
    \parbox[c]{0.22\textwidth}{\includegraphics[width=0.22\textwidth]{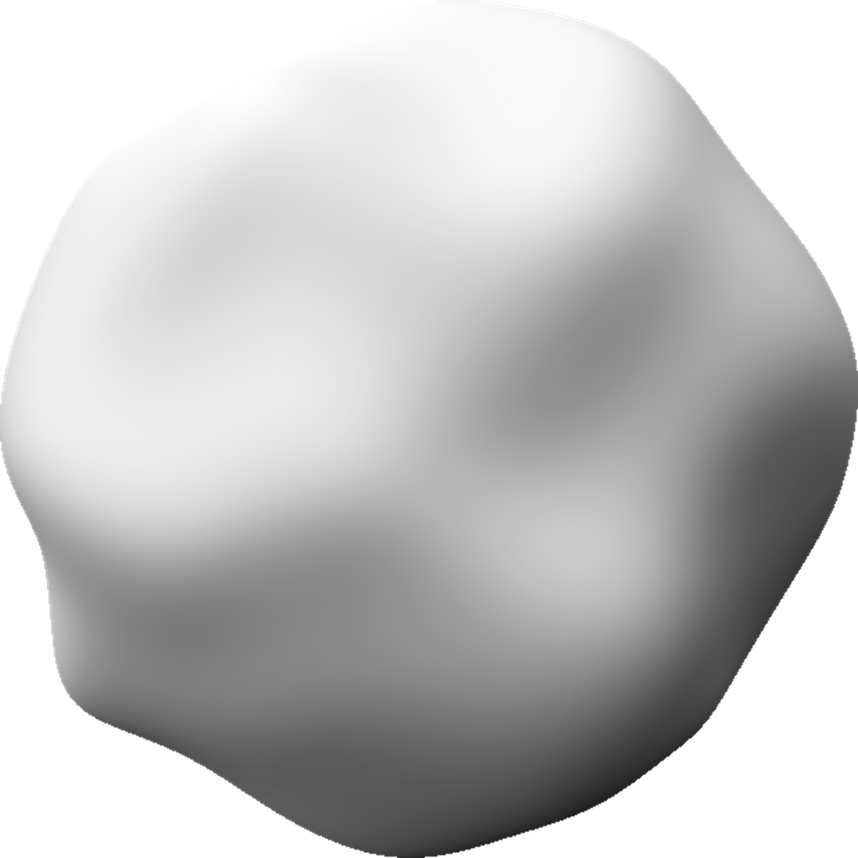}}
    &
    \parbox[c]{0.22\textwidth}{\includegraphics[width=0.22\textwidth]{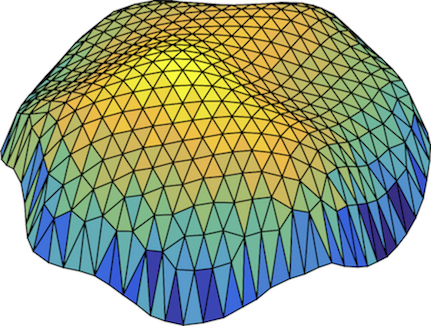}}
    &
    \parbox[c]{0.22\textwidth}{\includegraphics[width=0.22\textwidth]{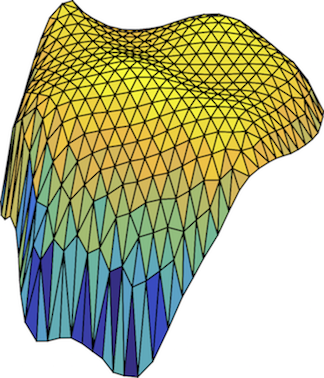}}
    &
    \parbox[c]{0.22\textwidth}{\includegraphics[width=0.22\textwidth]{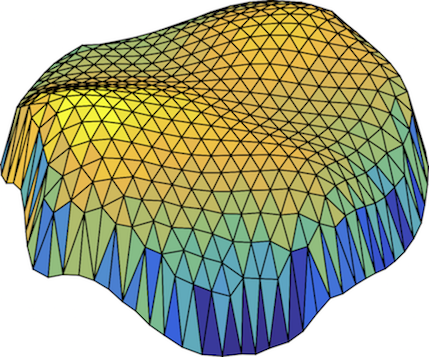}} \\
    \parbox[c]{0.22\textwidth}{\includegraphics[width=0.22\textwidth]{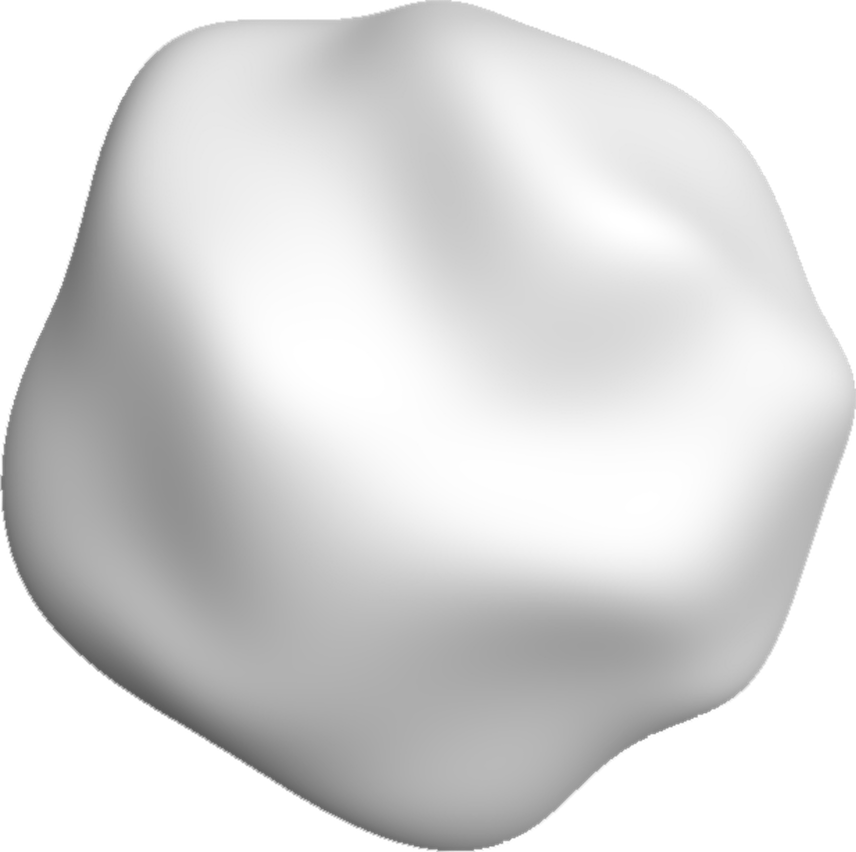}}
    &
    \parbox[c]{0.22\textwidth}{\includegraphics[width=0.22\textwidth]{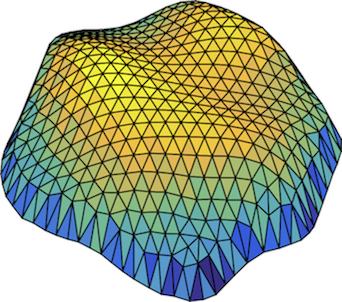}}
    &
    \parbox[c]{0.22\textwidth}{\includegraphics[width=0.22\textwidth]{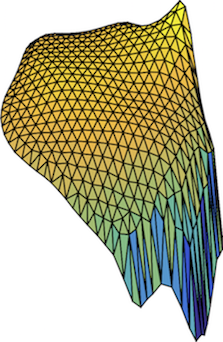}}
    &
    \parbox[c]{0.22\textwidth}{\includegraphics[width=0.22\textwidth]{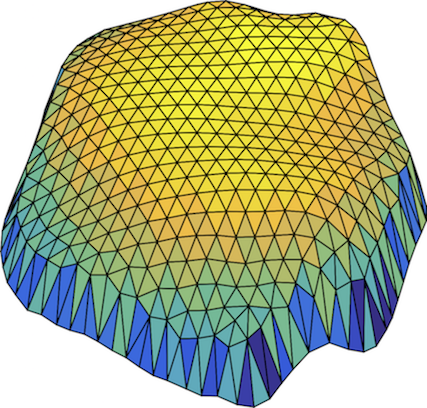}} \\
    \parbox[c]{0.22\textwidth}{\includegraphics[width=0.22\textwidth]{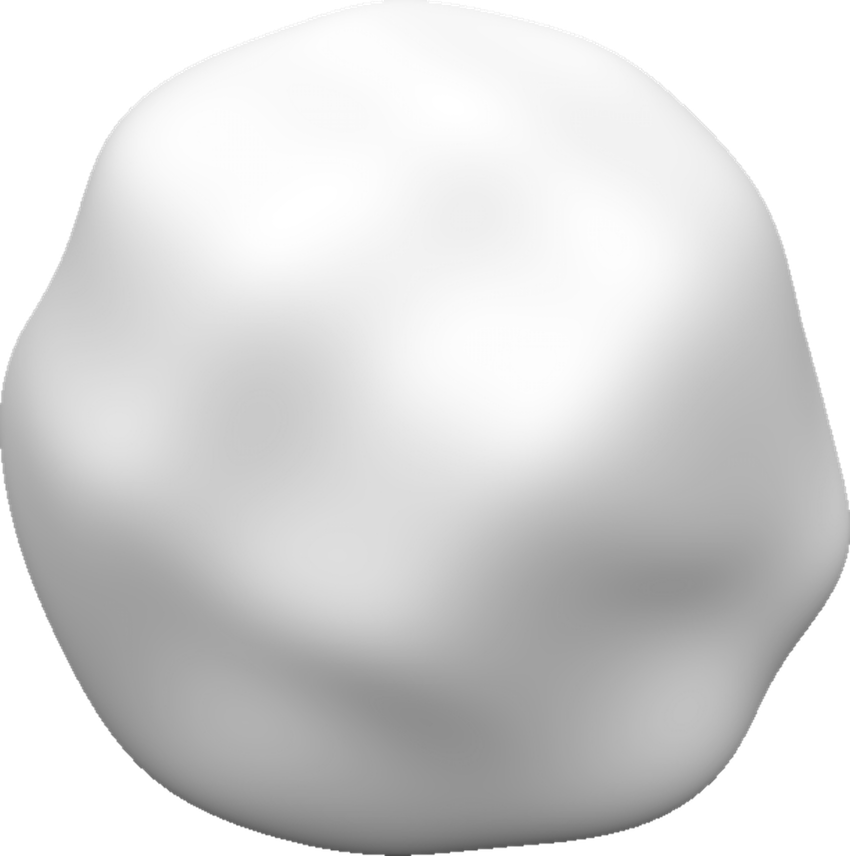}}
    &
    \parbox[c]{0.22\textwidth}{\includegraphics[width=0.22\textwidth]{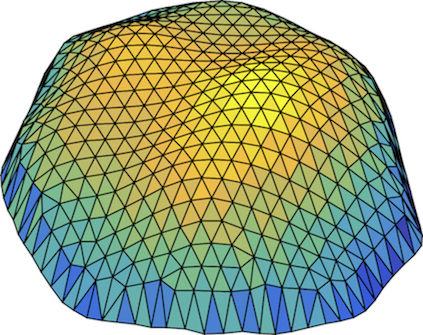}}
    &
    \parbox[c]{0.22\textwidth}{\includegraphics[width=0.22\textwidth]{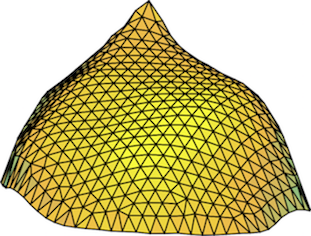}}
    &
    \parbox[c]{0.22\textwidth}{\includegraphics[width=0.22\textwidth]{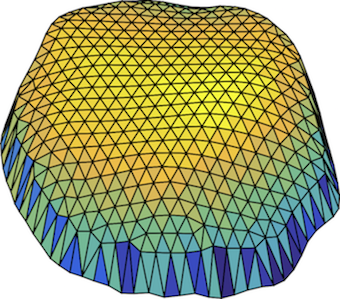}}
  \end{tabular}
  \caption{Example reconstructions, perturbed light source (shifted 22.5 degrees from true source direction).}
  \label{fig:pertrecs}
\end{table}

\vfill\eject
\begin{appendix}

\section{Mathematical Background}
\label{chapter:mathbackground}

In order to make this paper self-contained, we here introduce relevant background material as two Appendices. The idea is to provide a guide for the less experienced reader.
We begin with basic notions (and practical considerations) from differential geometry.  The main point is to illustrate the linear-algebraic structures that emerge when one considers (carefully) the different coordinate systems required. Key to understanding the main content in the paper is to appreciate how tensors arise after taking multiple derivatives. In the next Appendix, we review key ideas from tensor analysis. 

\subsection{Surfaces and Surface Normals}
\label{appendix:dg}

Since our main goal concerns three-dimensional shape, we begin by developing tools to analyze surfaces in $\R^3$. The material is standard and our goal is to show how derivatives lead to tensors.  For classical references see \cite{docarmo, oneill} and, especially, \cite{Dodson:1991uc}.

A \emph{parametrization} of a surface $S$ is given by a function $\vec{s(x,y)} : R \subset \R^2 \to S \subset \R^3$, taking points in a two-dimensional domain to points on the surface (embedded in a three dimensional ``ambient space'').

Taking partial derivatives of $\s$ with respect to the two parameters gives vectors in $\R^3$ that describe how surface position changes with changing position in the parameter domain. Specifically, fixing a point $\x_0 = (x_0,y_0)$ in the image, $\s_x(\x_0) = \frac{\partial \s}{\partial x}(\x_0)$ and $\s_y(\x_0) = \frac{\partial \s}{\partial y}(\x_0)$ are \emph{tangent vectors} to the surface at $\x_0$, and together span the \emph{tangent plane} of the surface at $\x_0$.

Stacking $\s_x(\x_0)$ and $\s_y(\x_0)$ side by side to form a matrix yields the $3\times2$ \emph{Jacobian matrix} of $\s$, \[
  \D\s|_{\x=\x_0} = \begin{pmatrix}\col{\s_x(\x_0)} & \col{\s_y(\x_0)}\end{pmatrix}.
\]
Since $\s$ is a map from the parameter space to the surface, $\D\s|_{\x_0}$ is a linear map from the tangent space associated with the point $\x_0$ in the parameter space (i.e., offsets from $\x_0$) to the tangent plane of the surface at $\s(\x_0)$. In other words, $\D\s$ translates ``steps in parameters'' to ``steps on the surface''.

In particular, a step $(u,v)$ in the parameters corresponds to a step $(u,v)$ on the surface with the same coordinates when expressed in the ``standard tangent basis'' given by the columns of $\D\s$. The corresponding $\R^3$ vector $\v$ can be recovered by expansion in this basis: $\v = \D\s \cdot (u, v)\tr$. (In the previous expression and subsequently we suppress reference to the point of evaluation $\x_0$---its presence should be implicitly assumed.)

Note that the standard tangent basis is not generally orthonormal---orthonormality occurs only when the surface is fronto-parallel at the point of evaluation. Therefore, to compute inner products between vectors in the tangent plane in a way compatible with inner products in the ambient space, we must expand into $\R^3$ and compute inner products there: \[
  \innerprod{\valpha}{\vbeta} = \innerprod{\D\s\,\valpha}{\D\s\,\vbeta} = \valpha\tr\,{\D\s}\tr \D\s\,\vbeta = \valpha\tr G\vbeta,
\]
where $G = {\D\s}\tr \D\s = \left(\!\begin{smallmatrix}\s_x\tr\s_x & \s_x\tr\s_y \\ \s_x\tr\s_y & \s_y\tr\s_y\end{smallmatrix}\!\right)$ is the matrix of inner products of the standard tangent basis vectors. $G$ is commonly known as the ``first fundamental form'' (often represented as $\mathrm{I}$, which we avoid due to potential confusion with the identity matrix). Computing inner products by multiplying against the first fundamental form means explicit expansion into $\R^3$ is unnecessary.

A common parameterization is the so-called ``Monge patch'' form, where $\s$ is given by $\s(x,y) = (x,y,h(x,y))\tr$. We adopt this parameterization in the material that follows. This allows one to think of the parameter space as the image plane, and $\s$ as a function taking points in the image to points on the surface. $\D\s$ then takes steps in the image to steps on the surface.

Just as $\s$ maps locations in the image to locations on the surface $S$, we define $\n : R \subset \R^2 \to S^2 : \x \mapsto \frac{\s_x(\x) \times \s_y(\x)}{\vecnorm{\s_x(\x) \times \s_y(\x)}}$ (where $S^2$ is the unit sphere in $\R^3$) to be the map taking a location $\x$ in the image to the surface normal at $\s(\x)$. It can be viewed as the composition $\nt \circ \s$, where $\nt : S \to S^2$ is the map taking points on the surface to the unit sphere, often referred to as the \emph{Gauss map}.

$\D\n = \D\nt|_{\s(\x_0)} \circ \D\s|_{\x_0}$ is the linear map expressing how the surface normal changes (at $\x_0$) with changing position in the image (the Jacobian matrix of $\n$). In the standard basis for $\R^3$, $\D\n$ is a $3 \times 2$ matrix. Since the normal is always unit length, we have $\n\tr\n = 1 \Rightarrow \n\tr\D\n = \vec{0}$, demonstrating that the column space of $\D\n$ (and $\D\nt$) is orthogonal to the normal and hence lies in the tangent plane.

This fact permits $2\times2$ matrix expressions for $\D\nt$ (the differential of the Gauss map) in any basis for the tangent plane. Commonly, $\D\nt$ is expressed in the standard tangent basis. We will denote this matrix $[\D\nt]_\beta^\beta$, where $\beta$ is used to indicate the standard tangent basis, and the subscript and superscript on the square brackets indicate (respectively) the bases used for inputs and outputs of the matrix. 

The eigenvectors of $\D\nt$ are called the \emph{principal directions}, and form the directions (in the tangent plane) of maximal and minimal normal change, while the eigenvalues are called the \emph{principal curvatures}, and express the (signed) magnitude of normal change in the corresponding principal directions. $\D\nt$ is also often referred to as the ``shape operator''. The product $G\,\D\nt$ -- the second fundamental form -- often is expressed $\mathrm{I\!I}$. The second fundamental form makes it easy to ``measure'' the amount of normal change in a given direction, since $\w\tr \mathrm{I\!I} \v = \langle \w, \D\nt(\v) \rangle$ is the inner product of $\w$ with the change in normal in the direction $\v$. When the basis for the tangent plane is orthonormal $G = I$ and the second fundamental form and $\D\nt$ are represented by the same matrix.

Higher derivatives of $\n$ are denoted $\D^2\n$, $\D^3\n$, etc., and form \emph{tensors} of progressively higher order. $\D\n$ (at some point $\x_0$) takes in one ``step'' in the image--say $\valpha$--and outputs (the first order approximation to) the corresponding change in the normal when moving (away from $\x_0$) with velocity $\valpha$. $\D^2\n$ takes in two directions in the image--say $\valpha$ and $\vbeta$---and outputs the change in [the change in the normal when moving with velocity $\valpha$] when moving with velocity $\vbeta$. In other words, $\D^2\n(\valpha, \vbeta)$ describes how the derivative of $\n$ in direction $\valpha$ changes in direction $\vbeta$. $\D^2\n$ is a multilinear map (linear in each of its inputs), and can be expressed in the standard basis for $\R^3$ as a $3\times2\times2$ ``third-order'' array of numbers, while $\D^3\n$ takes the form of a fourth-order $3\times2\times2\times2$ array, etc.

\subsection{Tensors}
\label{appendix:tensors}

There is a rich mathematical tradition linking tensors and differential geometry, which has been motivated by the theory of manifolds (classical references include \cite{bishop:gold, boothby:intro}; more recently, see e.g. \cite{landsberg:book}.  We especially recommend  \cite{Dodson:1991uc,} for the intuition it develops. Applications in physics have also been influential \cite{amp:c-b}, in particular
mechanics \cite{amr:manifolds} and general relativity \cite{MTW:book}. 
More recently, applications in signal processing have emerged \cite{comon:brief}.
The main use of tensors in the computer vision community 
is in multi-view and multi-camera stereo \cite{Ma:2003:IVI:971144}, which we do not discuss, and recently in medical imaging (diffusion MRI \cite{dti:principles, b:j:dti:review}). Several on-line introductions to this material are also available, e.g. \cite{florak:web, landsberg:web}.

Our emphasis is different. We have just seen how tensors arise naturally in the process of taking derivatives. We now review tensors and cover some related tools that we utilize when working with $\D^2\n$.
(Some care is required here, as this is only true for the right kind of derivative: when the basis used varies throughout the space under consideration, a ``covariant derivative'' that accounts for the change in basis from point to point is required \cite{}. The regular componentwise derivative and the covariant derivative coincide for spaces in which the basis is constant, such as $\R^n$ with the standard basis.) 

Although some definitions emphasize a view of tensors as multidimensional arrays having certain transformation properties under basis changes, we adopt the view of tensors as multilinear maps \cite{Dodson:1991uc}.  Specifically, a tensor $T$ is a map
\begin{equation}
  T : V_1^*\times \cdots \times V_m^*\times V_1\times \cdots \times V_n \to \R,
\end{equation}
where the $V_i$ are vector spaces and the $V_i^*$ are spaces of dual vectors (covectors)---linear functionals on a vector space, meaning they take vectors and return scalars. The number of vectors and covectors $T$ takes as input defines the \emph{order} of $T$. The number of vector inputs is the covariant order of $T$ ($n$ in the above definition), while the number of covector inputs determines the contravariant order ($m$ above), making $T$ an $(m,n)$ tensor (contravariant order first). Each different input ``slot'' is called a \emph{mode} of the tensor.

Every regular vector is a tensor of contravariant order 1 (and covariant order 0), and so can be considered as a linear functional on covectors (by taking the covector and applying it to the vector itself). Similarly, linear functionals are tensors of covariant order 1 (and contravariant order 0). We can define a \emph{tensor product} that ``glues together'' two tensors to form a higher order tensor by defining, for two tensors $T$ (of order $(m,n)\tr$) and $S$ (of order $(l,p)$), the $(m+l,n+p)$ order tensor
\begin{align*}
  &(T\otimes S)(\v^1,\ldots,\v^m,\v_1,\ldots,\v_n,\w^1,\ldots,\w^l,\w_1,\ldots,\w_p) \\
  &\qquad\qquad =
  T(\v^1,\ldots,\v^m,\v_1,\ldots,\v_n)S(\w^1,\ldots,\w^l,\w_1,\ldots,\w_p),
\end{align*}
which feeds each tensor its respective inputs and multiplies the results together.

Not all tensors are ``simple'' (or pure) tensors consisting only of tensor products of vectors and covectors. However, all tensors can be written as a \emph{sum} of such tensor products. The minimal required number of terms in the sum is known as the \emph{rank} of the tensor.

Choosing a basis for each vector and covector space, and forming all possible tensor products of basis vectors from each space, yields a basis for the space of tensors. Letting ${\b_i}_j$ represent the $j$'th basis vector for the vector space $V_i$ and ${\b_i}^j$ represent the $j$'th basis vector for the covector space $V^*_i$, $T$ above can be written as the sum
\begin{equation}
  T = \sum_{\substack{k_1,\ldots,k_m\\l_1,\ldots,l_n}}T^{k_1k_2\ldots k_m}_{l_1l_2\ldots l_n}{\b_1}_{k_1}\otimes \cdots\otimes{\b_m}_{k_m}\otimes{\b_1}^{l_1}\cdots\otimes{\b_n}^{l_n}
\end{equation}
where the upper indices correspond to the contravariant components and the lower indices to the covariant components (this is switched for the basis vectors, following \cite{Dodson:1991uc}---this permits easy use of Einstein notation, which we won't cover here). The scalars $T^{k_1k_2\ldots k_m}_{l_1l_2\ldots l_n}$ are precisely the elements of a multidimensional array representing the tensor in the chosen basis, with each index corresponding to elements along a different mode (with length the dimension of the corresponding vector space) of the array.

If a tensor has two modes whose corresponding vector spaces are dual to one another (a $V$ and $V^*$), these modes can be ``contracted''. If the tensor has the form (a sum of simple tensors) $T = \sum_i\cdots\otimes\v^*_i\otimes\cdots\otimes\v^i\otimes\cdots$, the contraction of the two modes is given by $T' = \sum_i\cdots\otimes\v^*_i(\v^i)\otimes\cdots$, applying the linear functionals $\v_i^*\in V^*$ to the corresponding vectors $\v^i\in V$. If $T$ is order $(m,n)$, then $T'$ is order $(m-1,n-1)$.

This permits a view of a tensor $T$ as a linear functional on the space of tensors where each vector and covector space associated with $T$ has been replaced by its dual. To evaluate $T$ against a tensor $T'$ in this dual tensor space, we can simply form the tensor product $T\otimes T'$ and contract all the corresponding modes to yield a scalar.

The contraction operation is a generalization of the matrix trace. A matrix can be seen as a $(1,1)$ tensor, and via the SVD can be decomposed into the sum of outer products of row and column vectors:
\begin{equation*}
   M = U\Sigma V\tr = \sum_i \sigma_i\vec{u}_i \v_i\tr,
 \end{equation*}
Contraction is then given by
\begin{equation*}
  \sum_i \sigma_i\v_i\tr\vec{u}_i = \sum_i \sigma_i\trace(\v_i\tr\vec{u}_i) = \sum_i \sigma_i\trace(\vec{u}_i\v\tr) = \trace(M).
\end{equation*}

If $T'$ is a tensor formed by (sums of) tensor products of vectors from only \emph{some} of the duals to $T$'s associated vector spaces, we can view $T$ as a linear map, applying it to $T'$ by a tensor product followed by contractions on the relevant modes, and yielding another tensor formed by the ``leftover'' modes of $T$ that weren't involved in the contractions.

This view is particularly fruitful, because it suggests there should be a matrix representation for $T$ (in addition to its representation as a multidimensional array). Indeed there is! To make the presentation easier, we forego generality and consider only a third-order (3-mode) tensor, $T \in W \otimes V^* \otimes V^*$ (or as a multilinear map $T : W^* \times V \times V \to \R$). By the above we can view $T$ as a linear map $T : V \otimes V \to W$. If $V$ is two-dimensional, say, then elements of $V \otimes V$, considered as vectors, have 4 elements. If $W$ is three-dimensional, $T$ can thus be viewed as a $3\times4$ matrix.

The matrix version of $T$ (for a specific linear map ``perspective'') is related to $T$'s $3\times2\times2$ multidimensional array by an unfolding process. Let $T$'s first mode run down the page, second mode run across, and third mode go ``into'' the page. Then the four vertical columns of $T$, when stacked horizontally, form an ``unfolding'' of $T$ into a matrix. (Other unfoldings are also possible, by stacking the rows or ``depths'' side by side as columns.)

This notion is very useful, because it allows tools from linear algebra developed for matrices---such as the SVD---to be applied in a principled way to tensors. This is utilized heavily in \cite{Lathauwer:2000}, which develops a higher-order analog of the SVD for tensors by considering the SVD's of different unfoldings of a tensor.

How are the ``vectorized'' representations of elements of $V \otimes V$ formed? In the context of the current example, these are 4-element column vectors. Given $\v, \v' \in V$, define
\begin{equation}
  \v \otimes \v' = \begin{pmatrix}
    v^1 \v' \\
    v^2 \v'
  \end{pmatrix},
\end{equation}
where $v^1$ and $v^2$ are the components of $\v$. This is a special case of the \emph{Kronecker product}. If we have two matrices $A$, $B$, representing linear maps from $V$ to $V$, we can combine them to give a linear map from $V\otimes V$ to $V\otimes V$ by constructing the matrix
\begin{equation}
  A\otimes B = \begin{pmatrix}
    a_{11} B & a_{12} B \\
    a_{21} B & a_{22} B
  \end{pmatrix}.
\end{equation}
The Kronecker product has the intuitive property that $(A\otimes B)(\v\otimes\v') = (A\v)\otimes(B\v')$.

The above tools, especially the unfolding operation and the Kronecker product, will be applied to analysis of $\D^2\n$ in \ref{chapter:stats}.


\section{Decomposition of {$\D\n$}}
\label{sec:dndecomp}

In working with derivatives of the normal, it is common to use $[\D\nt]_\beta^\beta$, the differential of the Gauss map expressed in the standard tangent basis. A principle reason for use of the standard tangent basis is that vectors in this basis relate in a straightforward manner to directions in the image, since they have the same component representations. However, we find it easier to work with and reason about $[\D\n]_{\epsilon_2}^{\epsilon_3}$, considered as a map from directions in the image to normal changes in $\R^3$. In particular, this simplifies analysis of image derivatives, since they don't require explicit projection of the light source into the tangent plane (and for $\D^2\n$, subsequent covariant differentiation).

Observe that $[\D\n]_{\epsilon_2}^{\epsilon_3}$ it can be written as the following composition: \[
  [\D\n]_{\epsilon_2}^{\epsilon_3} = [I]_\beta^{\epsilon_3}\,[\D\nt]_\beta^\beta\,[\D\s]_{\epsilon_2}^\beta
\]
where $\D\s$ is the differential of the parametrization $\s$, and the sub- and superscripts on the brackets respectively signify the input and output bases of the associated matrices: $\epsilon_2$ and $\epsilon_3$ are the standard bases for $\R^2$ (the image) and $\R^3$ (the ambient space), while $\beta$ is the standard tangent basis formed by the columns of $\D\s$. Thus $[\D\s]_{\epsilon_2}^\beta$ takes a step in the image and expresses it in the standard tangent basis, $[\D\nt]_\beta^\beta$ calculates the change in normal associated with this step in the tangent plane (expressing the output once again in the standard tangent basis), and $[I]_\beta^{\epsilon_3}$ expands the result as a vector in $\R^3$.

However, there is no reason we are restricted to the choice of basis $\beta$ above. The $\beta$ basis is not orthonormal, which means that $[\D\nt]_\beta^\beta$, while diagonalizable (with the principal curvatures as eigenvalues), does not have orthonormal eigenvectors unless the appropriate inner product (the first fundamental form $G$) is used. Specifically, we have $[\D\nt]_\beta^\beta = \Wt K \Wt^{-1}$ with $\Wt\tr G \Wt = I \ne \Wt\tr\Wt$, where $K$ is the diagonal matrix of principal curvatures and $\Wt$ contains as its columns the principal directions expressed in the standard tangent basis.

To make subsequent analysis easier, we seek an appropriate orthogonal basis for the tangent plane. Any orthonormal basis would do, but we show that two are particularly natural. More formally, we want a basis transformation $B$ such that $W\tr W = (B\Wt)\tr(B\Wt) = \Wt\tr\!B\tr\! B \Wt = I$. Since $\Wt\tr G\Wt = I$ also, this suggests finding $B\tr\!B = G$. With $\D\s = \U0\S0\V0\tr$ being the SVD of $\D\s$, $G = \D\s\tr\D\s = \V0\S0^2\V0\tr$, so we have two obvious choices for $B$, $B = \V0\S0\V0\tr$, or $B = \S0\V0\tr$ (additionally, any $R\Sigma V\tr$ for $R$ orthogonal would work). Choosing $B = \V0\S0\V0\tr$ yields a basis in the tangent plane that is a rotation of the $x$-$y$ standard basis in the image around the direction in the image plane orthogonal to the tilt direction of the surface. Choosing $B = \S0\V0\tr$, on the other hand, yields a basis formed by the tilt (surface gradient) direction and its perpendicular in the tangent plane. The distinction between these choices is minor, and amounts to whether the principal directions and tilt are specified independently, or whether principal directions are specified relative to the tilt direction.

For brevity, we adopt the choice $B = \S0\V0\tr$. Call the associated basis $\gamma$. Then $[I]_\gamma^{\epsilon_3} = U$, since the columns of $U$ are precisely the tilt direction and its perpendicular in the tangent plane; $[\D\nt]_\gamma^\gamma = WKW\tr$, where the columns of $W$ are the principal directions and $K$ is diagonal; and $[\D\s]_{\epsilon_2}^\gamma = \Sigma V\tr$, which rotates image vectors into the tilt basis and scales them to account for foreshortening. This yields the decomposition:
\begin{align*}
  \D\n &= \U0 W K W\tr\S0\V0\tr \label{eq:dn-decomp} \yestag \\
  \D\n^+ &= \V0\S0^{-1} W K^{-1} W\tr U\tr. \label{eq:dn+-decomp} \yestag
\end{align*}

\subsection{Interpretation as a Taylor Expansion from the Tangent Plane}
\label{sec:dntaylor}

As another perspective on the decomposition of $\D\n$ above that facilitates extending the decomposition to third order, imagine the surface is fronto-parallel. Then a second-order Taylor expansion of the surface is given by
\begin{align}
  h(x,y)
  &= \frac{1}{2}h_{xx}(0,0)\, x^2 + h_{xy}(0,0)\, xy + \frac{1}{2}h_{yy}(0,0)\, y^2 \\
  &= \frac{1}{2}\x\tr H \x,
\end{align}
where $H$ is the surface Hessian and $\x = (x,y)\tr$. The partial derivatives $h_x$ and $h_y$ at $\x$ are given by $(h_x(\x), h_y(\x))\tr = H\x$, so letting $n_3(\x) = \frac{1}{\sqrt{1 + h_x(\x)^2 + h_y(\x)^2}}$ be the normalizing factor, the normal is
\begin{equation}
  \n(\x) = n_3(\x)\begin{pmatrix}
    h_x(\x) \\ h_y(\x) \\ 1
  \end{pmatrix}
  =
  n_3(\x)\begin{pmatrix}
    H\x \\ 1
  \end{pmatrix}.
\end{equation}
Then
\begin{align*}
  \D\n|_{\x} &= \begin{pmatrix}
    H\x \\ 1
  \end{pmatrix}\pd{n_3(\x)}{\x}
  +
  n_3(\x)\begin{pmatrix}
    \multicolumn{2}{c}H \\ 0 & 0
  \end{pmatrix} \yestag \label{eq:dnatx} \\
  &\alignedarrow \\
  \D\n|_{\x = 0} &= \begin{pmatrix}
    \multicolumn{2}{c}H \\ 0 & 0
  \end{pmatrix},
\end{align*}
because $\n(0) = \zhat$, the $z$-axis vector, and $\n\tr\D\n = 0$ implies $\pd{n_3(\x)}{\x} = 0$.

The decomposition above can thus be seen as taking a second order Taylor expansion ``from'' the principal curvature directions basis in the tangent plane, and rotating into (from the image) and out of (into $\R^3$) this basis appropriately. In this basis, $h_{xx}(0,0)$ an $h_{yy}(0,0)$ are precisely the principal curvatures (and $h_{xy}(0,0) = 0$). This perspective extends nicely to handling second derivatives of the normal. 

\section{Decomposition of {$\D^2\n$}}
\label{sec:d2ndecomp}

Following the Taylor expansion view above (\ref{sec:dntaylor}), we take a third order Taylor expansion of the surface (imagining it was fronto-parallel and the $x$ and $y$ axes align with the principal curvature directions):
\begin{align*}
  h(x,y)
  &= \frac{1}{2}\left(\kappa_1 x^2 + \kappa_2 y^2\right) + \frac{1}{6}\left(f x^3 + 3g x^2 y + 3h x y^2 + i y^3\right) \\
  &= \frac{1}{2}\x\tr K\x + \frac{1}{6}\calK(\x,\x,\x),
\end{align*}
where $\calK$ is the $2\times 2 \times 2$ symmetric tensor
\begin{equation}
  \calK = \left(\begin{array}{@{}cc@{\enskip}|@{\enskip}cc@{}}
    f & g & g & h \\
    g & h & h & i
  \end{array}\right),
\end{equation}
with the vertical bar separating the front and back ``planes'' of the array. $\calK(\x,\x,\x)$ applies $\calK$ to three copies of $\x$ via linear combinations along each of the modes of $\calK$. We note that $f = \kf$, $g = \kg$, $h = \kh$, $i = \ki$, the partial derivatives of the principal curvatures.

Then as above (with mild abuse of notation concerning the ``three-dimensional'' nature of some terms below)
\begin{gather*}
  \n(\x) = n_3(\x)\begin{pmatrix}
    K\x + \frac{1}{2}\calK(I,\x,\x) \\ 1
  \end{pmatrix} \displaybreak[0]\\
  \Longrightarrow \\
  \D\n|_{\x} = \begin{pmatrix}
    K\x + \frac{1}{2}\calK(I,\x,\x) \\ 1
  \end{pmatrix}\pd{n_3(\x)}{\x}
  +
  n_3(\x)\begin{pmatrix}
    \multicolumn{2}{c}{K + \calK(I,I,\x)} \\ \multicolumn{2}{c}{0 \qquad 0}
  \end{pmatrix} \displaybreak[0]\\
  \Longrightarrow \\
  \begin{aligned}
    \D^2\n|_{\x}
    &= \begin{pmatrix}
      \multicolumn{2}{c}{K + \calK(I,I,\x)} \\ \multicolumn{2}{c}{0 \qquad 0}
    \end{pmatrix} \pd{n_3(\x)}{\x} \\
    &+ \begin{pmatrix}
      K\x + \frac{1}{2}\calK(I,\x,\x) \\ 1
    \end{pmatrix}\pdd{n_3(\x)}{\x} \\
    &+ \pd{n_3(\x)}{\x} \begin{pmatrix}
    \multicolumn{2}{c}{K + \calK(I,I,\x)} \\ \multicolumn{2}{c}{0 \qquad 0}
  \end{pmatrix} \\
    &+ n_3(\x)\begin{pmatrix}
      \calK \\
      0 \; 0 \; | \; 0 \; 0
    \end{pmatrix}.
  \end{aligned}
\end{gather*}
Evaluating at $\x = 0$ gives
\begin{align*}
  \D^2\n|_{\x = 0}
  &=
  \begin{pmatrix}
    0 \\ 0 \\ 1
  \end{pmatrix}\pdd{n_3(\x)}{\x}
  +
  \begin{pmatrix}
    \calK \\
    0 \; 0 \; | \; 0 \; 0
  \end{pmatrix} \displaybreak[0]\\
  &=
  \begin{pmatrix}
    \calK \\
    \kappa_1^2 \; 0 \; | \; 0 \; \kappa_2^2
  \end{pmatrix} \\
  &=
  \calA,
\end{align*}
where the evaluation of $\pdd{n_3(\x)}{\x}$ can be seen by noting $n_3(\x) = \frac{1}{\sqrt{1 + \x\tr K^2 \x}}$, which after two derivatives and evaluation at 0 leaves only $K^2$ in the numerator.

Unfolding $\calA$ to give $\calA_{(1)}$ (the ``mode-1'' unfolding) by stacking the columns side by side, and using the Kronecker product, yields the final decomposition for general (non-fronto-parallel) $\D^2\n$ as a $3\times 4$ unfolded matrix, as before rotating ``into'' and ``out of'' the principal curvatures basis:
\begin{equation}
  \D^2\n = U_3W_3\calA_{(1)}(W\tr\Sigma V\tr)^{\otimes2}.
\end{equation}
We use the notation $U_3$ and $W_3$ to denote the extensions of the $U$ and $W$ matrices to orthogonal $3\times 3$ forms---for $U$, this involves addition of the normal as a third column, while $W_3$ embeds $W$ in the upper left $2\times 2$ submatrix of a $3\times 3$ identity matrix.

The pseudoinverse is given by
\begin{equation}
  \D^2\n^+_{(1)}
  =
  (\V0\S0^{-1} W)^{\otimes 2}\calAunf^+W\tr\U0\tr,
\end{equation}
To compute $\calA^+$, instead of evaluating the pseudo-inverse via the SVD, we can use the more explicit form \[
  \calAunf^+ = \calAunf\tr\left(\calAunf\calAunf\tr\right)^{-1},
\]
valid when $\calAunf$ has full row rank. We evaluated this somewhat daunting expression using a computer algebra system, which yields \[
  \calAunf^+ = \frac{1}{m} \begin{pmatrix}
    -h \kappa _2^2 & g \kappa _2^2 & h^2-g i \\
    \frac{1}{2} \left(g \kappa _2^2-i \kappa _1^2\right) & \frac{1}{2} \left(h \kappa _1^2-f \kappa _2^2\right) & \frac{1}{2} (f i-g h) \\
    \frac{1}{2} \left(g \kappa _2^2-i \kappa _1^2\right) & \frac{1}{2} \left(h \kappa _1^2-f \kappa _2^2\right) & \frac{1}{2} (f i-g h) \\
    h \kappa _1^2 & -g \kappa _1^2 & g^2-f h \\
  \end{pmatrix},
\]
where $m = \kappa_1^2\left(h^2-g i\right) + \kappa_2^2\left(g^2-f h\right)$.

\end{appendix}

\bibliography{Thesis}

\begin{thebibliography}{10}

\bibitem{amr:manifolds}
R.~Abraham, J.E. Marsden, and T.~Ratiu.
\newblock {\em Manifolds, Tensor Analysis and Applications}.
\newblock Academic Press, 1975.

\bibitem{Barron:2012tt}
Jonathan~T Barron and Jitendra Malik.
\newblock {Shape, albedo, and illumination from a single image of an unknown
  object}.
\newblock In {\em IEEE Conference on Computer Vision and Pattern Recognition},
  pages 334--341. IEEE, 2012.

\bibitem{b:j:dti:review}
P.J. Basser and D.K. Jones.
\newblock Diffusion-tensor mri: Theory, experimental design and data analysis -
  a technical review.
\newblock {\em NMR in Biomedicine}, 15(7-8):456--467--539, 2002.

\bibitem{Belhumeur:1999hd}
Peter~N Belhumeur, David~J Kriegman, and Alan~L Yuille.
\newblock {The Bas-Relief Ambiguity}.
\newblock {\em International Journal Of Computer Vision}, 35(1):33--44, 1999.

\bibitem{ben2004geometrical}
Ohad Ben-Shahar and Steven Zucker.
\newblock Geometrical computations explain projection patterns of long-range
  horizontal connections in visual cortex.
\newblock {\em Neural computation}, 16(3):445--476, 2004.

\bibitem{bishop:gold}
Richard~L. Bishop and Samuel~I. Goldberg.
\newblock {\em Tensor Analysis on Manifolds}.
\newblock Dover Publications, New York, 1968.

\bibitem{boothby:intro}
William~M. Boothby.
\newblock {\em An Introduction to Differentiable Manifolds and Riemannian
  Geometry}.
\newblock Academic Press, 1975.

\bibitem{Breton:1996ic}
Pierre Breton and Steven~W Zucker.
\newblock {Shadows and shading flow fields}.
\newblock {\em IEEE Conference on Computer Vision and Pattern Recognition},
  pages 782--789, 1996.

\bibitem{Chen:2000ba}
H~F Chen, Peter~N Belhumeur, and David~W Jacobs.
\newblock {In search of illumination invariants}.
\newblock {\em IEEE Conference on Computer Vision and Pattern Recognition},
  1:254--261 vol.1, 2000.

\bibitem{amp:c-b}
Yvonne Choquet-Bruhat and Cecile DeWitt-Morette.
\newblock {\em Analysis, Manifolds and Physics}.
\newblock Elsevier Science Publishing Co., 1977.

\bibitem{comon:brief}
P.~Comon.
\newblock Tensors : A brief introduction.
\newblock {\em IEEE Signal Processing Magazine}, 31(3):44--53, May 2014.

\bibitem{Lathauwer:2000}
Lieven De~Lathauwer, Bart De~Moor, and Joos Vandewalle.
\newblock {A Multilinear Singular Value Decomposition}.
\newblock {\em SIAM Journal on Matrix Analysis and Applications}, 21(4), March
  2000.

\bibitem{Dieft81}
P.~Dieft and J.~Sylvester.
\newblock Some remarks on the shape-from-shading problem in computer vision.
\newblock {\em Journal of Mathematical Analysis and Applications},
  84(1):235--248, 1981.

\bibitem{docarmo}
M.P. Docarmo.
\newblock {\em Differential {G}eometry of {C}urves and {S}urfaces}.
\newblock Prentice-Hall Inc., Upper Saddle River, New Jersey, 1976.

\bibitem{Dodson:1991uc}
C~T~J Dodson and Timothy Poston.
\newblock {\em {Tensor Geometry}}.
\newblock The Geometric Viewpoint and Its Uses. Springer Science {\&} Business
  Media, 1991.

\bibitem{Ecker:2010uh}
A.~Ecker and A.~D. Jepson.
\newblock Polynomial shape from shading.
\newblock In {\em 2010 IEEE Computer Society Conference on Computer Vision and
  Pattern Recognition}, pages 145--152, June 2010.

\bibitem{Erens:1993vh}
R~G Erens, A~M Kappers, and Jan~J Koenderink.
\newblock {Perception of local shape from shading.}
\newblock {\em Perception {\&} psychophysics}, 54(2):145--156, August 1993.

\bibitem{Ferrie:1992vq}
F~P Ferrie and J~Lagarde.
\newblock {Curvature Consistency Improves Local Shading Analysis}.
\newblock {\em CVGIP: Image Understanding}, 55(1):95--105, January 1992.

\bibitem{Fleming:2013gx}
R~Fleming, R~Vergne, and S~Zucker.
\newblock {Predicting the effects of illumination in shape from shading}.
\newblock {\em Journal of Vision}, 13(9):611--611, 2013.

\bibitem{Fleming:2011dd}
Roland~W Fleming, Daniel Holtmann-Rice, and Heinrich~H B{\"u}lthoff.
\newblock {Estimation of 3D shape from image orientations}.
\newblock {\em Proceedings of the National Academy of Sciences of the United
  States of America}, 108(51):20438--20443, December 2011.

\bibitem{Fleming:2004bj}
Roland~W Fleming, Antonio Torralba, and Edward~H Adelson.
\newblock {Specular reflections and the perception of shape}.
\newblock {\em Journal of Vision}, 4(9):798--820, September 2004.

\bibitem{florak:web}
Luc Florak.
\newblock {\em Tensor Calculus and Differential Geometry}, 2016 (accessed
  February 9, 2017).

\bibitem{Freeman:1994br}
William~T Freeman.
\newblock {The generic viewpoint assumption in a framework for visual
  perception.}
\newblock {\em Nature}, 368(6471):542--545, April 1994.

\bibitem{Freeman:1996jc}
William~T Freeman.
\newblock {Exploiting the generic viewpoint assumption}.
\newblock {\em International Journal Of Computer Vision}, 20(3):243--261, 1996.

\bibitem{Garding:1993gz}
J.~G{\aa}rding.
\newblock {Direct estimation of shape from texture}.
\newblock {\em Pattern Analysis and Machine Intelligence, IEEE Transactions
  on}, 15(11):1202--1208, 1993.

\bibitem{Horn:1970tf}
Berthold K~P Horn.
\newblock {\em {Shape from shading: a method for obtaining the shape of a
  smooth opaque object from one view}}.
\newblock PhD thesis, Massachusetts Institute of Technology, 1970.

\bibitem{Horn:1975tq}
Berthold K~P Horn.
\newblock {Obtaining shape from shading information}.
\newblock In {\em The psychology of computer vision}, pages 115--155.
  McGraw-Hill, New York, 1975.

\bibitem{Horn:1977vn}
Berthold K~P Horn.
\newblock {Understanding image intensities}.
\newblock {\em Artificial Intelligence}, 1977.

\bibitem{Ikeuchi:1981dq}
K~Ikeuchi and Berthold K~P Horn.
\newblock {Numerical Shape From Shading and Occluding Boundaries}.
\newblock {\em Artificial Intelligence}, 17(1-3):141--184, 1981.

\bibitem{Judd:2007hm}
Tilke Judd, Fr{\'e}do Durand, and Edward~H Adelson.
\newblock {Apparent ridges for line drawing}.
\newblock {\em ACM Trans. Graph.}, 26(3):19:1--7, August 2007.

\bibitem{Knill:2014hp}
O~Knill.
\newblock {Cauchy{\textendash}Binet for pseudo-determinants}.
\newblock {\em Linear Algebra and its Applications}, 459:522--547, 2014.

\bibitem{Koenderink:1980bm}
Jan~J Koenderink and Andrea~J van Doorn.
\newblock {Photometric invariants related to solid shape}.
\newblock {\em Journal of Modern Optics}, 27(7):981--996, 1980.

\bibitem{tensors-arxiv-2}
Benjamin Kunsberg, Daniel Holtman-Rice, and Steven~W. Zucker.
\newblock {What's In A Patch, II: Visualizing generic surfaces}.
\newblock {\em ArXiv e-prints}, 2017.

\bibitem{Kunsberg:2014gua}
Benjamin Kunsberg and Steven~W Zucker.
\newblock {How Shading Constrains Surface Patches without Knowledge of Light
  Sources}.
\newblock {\em SIAM Journal on Imaging Sciences}, 7(2):641--668, April 2014.

\bibitem{landsberg:web}
J.M. Landsberg.
\newblock {\em Tensors: Geometry and Applications}, 2009 (accessed February 9,
  2017).

\bibitem{landsberg:book}
J.M. Landsberg.
\newblock {\em Tensors: Geometry and Applications}.
\newblock Graduate Studies in Mathematics. American Mathematical Society, 2012.

\bibitem{Liu:2004gm}
Baoxia Liu and James~T Todd.
\newblock {Perceptual biases in the interpretation of 3D shape from shading.}
\newblock {\em Vision Research}, 44(18):2135--2145, 2004.

\bibitem{Liua:1989vk}
D~C Liua and J~Nocedal.
\newblock {\em {On the limited memory BFGS method for large scale optimization
  problems}}.
\newblock Math. Program, 1989.

\bibitem{Ma:2003:IVI:971144}
Yi~Ma, Stefano Soatto, Jana Kosecka, and S.~Shankar Sastry.
\newblock {\em An Invitation to 3-D Vision: From Images to Geometric Models}.
\newblock SpringerVerlag, 2003.

\bibitem{MTW:book}
Charles~W. Misner, Kip~S. Thorne, and John~Archibald Wheeler.
\newblock {\em {Gravitation}}.
\newblock W. H. Freeman, 1973.

\bibitem{dti:principles}
Susumu Mori and Jiangyang Zhang.
\newblock Principles of diffusion tensor imaging and its applications to basic
  neuroscience research.
\newblock {\em Neuron}, 51(5):527--539, 2006.

\bibitem{oneill}
Barrett O'Neill.
\newblock {\em Elementary {D}ifferential {G}eometry, Revised 2nd Edition}.
\newblock Elsevier, Burlington, Massachusetts, 2006.

\bibitem{OShea:2010cp}
James~P O'Shea, Maneesh Agrawala, and Martin~S Banks.
\newblock {The influence of shape cues on the perception of lighting
  direction.}
\newblock {\em Journal of Vision}, 10(12):21, 2010.

\bibitem{Pentland:1984vy}
Alex~P Pentland.
\newblock {Local shading analysis.}
\newblock {\em Pattern Analysis and Machine Intelligence, IEEE Transactions
  on}, 6(2):170--187, February 1984.

\bibitem{Pentland:1990hv}
Alex~P Pentland.
\newblock {Linear shape from shading}.
\newblock {\em International Journal Of Computer Vision}, 4(2):153--162, 1990.

\bibitem{Prados:2005ca}
E~Prados and Olivier Faugeras.
\newblock {Shape from shading: a well-posed problem?}
\newblock {\em Computer Vision and Pattern Recognition}, 2:870--877, 2005.

\bibitem{Prados:2006gu}
Emmanuel Prados, Fabio Camilli, and Olivier Faugeras.
\newblock {A Unifying and Rigorous Shape from Shading Method Adapted to
  Realistic Data and Applications}.
\newblock {\em Journal of Mathematical Imaging and Vision}, 25(3):307--328,
  2006.

\bibitem{Rusinkiewicz:2004ia}
Szymon Rusinkiewicz.
\newblock {Estimating curvatures and their derivatives on triangle meshes}.
\newblock In {\em Symposium on 3D Data Processing, Visualization and
  Transmission}, pages 486--493, 2004.

\bibitem{sarti2008symplectic}
Alessandro Sarti, Giovanna Citti, and Jean Petitot.
\newblock The symplectic structure of the primary visual cortex.
\newblock {\em Biological Cybernetics}, 98(1):33--48, 2008.

\bibitem{2012arXiv1206.6445T}
Y.~{Tang}, R.~{Salakhutdinov}, and G.~{Hinton}.
\newblock {Deep Lambertian Networks}.
\newblock {\em ArXiv e-prints}, June 2012.

\bibitem{Todd:1983vv}
James~T Todd and E~Mingolla.
\newblock {Perception of surface curvature and direction of illumination from
  patterns of shading.}
\newblock {\em Journal of experimental psychology Human perception and
  performance}, 9(4):583--595, August 1983.

\bibitem{Wang:2013fq}
C~Wang, N~Komodakis, and N~Paragios.
\newblock {Markov random field modeling, inference {\&} learning in computer
  vision {\&} image understanding: A survey}.
\newblock {\em Computer Vision and Image Understanding}, 117(11):1610--1627,
  2013.

\bibitem{Worthington:1999um}
P~L Worthington and E~R Hancock.
\newblock {New constraints on data-closeness and needle map consistency for
  shape-from-shading}.
\newblock {\em Pattern Analysis and Machine Intelligence, IEEE Transactions
  on}, 21(12):1250--1267, December 1999.

\bibitem{Xiong:2015hj}
Ying Xiong, Ayan Chakrabarti, Ronen Basri, Steven~J Gortler, David~W Jacobs,
  and Todd Zickler.
\newblock {From Shading to Local Shape}.
\newblock {\em Pattern Analysis and Machine Intelligence, IEEE Transactions
  on}, 37(1):67--79, January 2015.

\bibitem{Connor08}
Yukako Yamane, {Eric T.} Carlson, {Katherine C.} Bowman, Zhihong Wang, and
  {Charles E.} Connor.
\newblock A neural code for three-dimensional object shape in macaque
  inferotemporal cortex.
\newblock {\em Nature Neuroscience}, 11(11):1352--1360, 11 2008.

\bibitem{Zhang:1999wm}
R~Zhang, P~S Tsai, J~E Cryer, and M~Shah.
\newblock {Shape from shading: A survey}.
\newblock {\em Pattern Analysis and Machine Intelligence, IEEE Transactions
  on}, 21(8):690--706, August 1999.

\bibitem{Zheng:1991tg}
Qinfen Zheng and Rama Chellappa.
\newblock {Estimation of Illuminant Direction, Albedo, and Shape from Shading}.
\newblock {\em IEEE Transactions on Pattern Analysis and Machine Intelligence},
  13:37, 1991.

\bibitem{Zoran:2014to}
Daniel Zoran, Dilip Krishnan, Jose Bento, and William~T Freeman.
\newblock {Shape and Illumination from Shading using the Generic Viewpoint
  Assumption}.
\newblock In {\em Advances in Neural Information Processing Systems}, 2014.

\end{thebibliography}

\end{document}